\documentclass{article} %
\usepackage{iclr2027_conference,times}
\usepackage{booktabs}
\usepackage{array}
\usepackage{tabularx}
\usepackage{enumitem}
\usepackage{tikz}
\usetikzlibrary{arrows.meta,positioning,calc,fit}
\usepackage{microtype}
\usepackage{caption}
\usepackage{amsthm}
\usepackage{amssymb}
\usepackage{amsmath}
\usepackage{mathtools}
\usepackage{wrapfig}
\usepackage{adjustbox}
\usepackage{float}
\usepackage{multirow}
\usepackage[table]{xcolor}
\usepackage{arydshln}

\usepackage{amsmath,amsfonts,bm}

\def\eqref#1{equation~\ref{#1}}

\def\1{\bm{1}}

\DeclareMathAlphabet{\mathsfit}{\encodingdefault}{\sfdefault}{m}{sl}
\SetMathAlphabet{\mathsfit}{bold}{\encodingdefault}{\sfdefault}{bx}{n}

\usepackage{hyperref}
\usepackage{xurl}

\definecolor{myblack}{RGB}{43,78,134}
\definecolor{mygray}{RGB}{245,246,248}
\definecolor{mygreen}{RGB}{38,120,84}
\definecolor{myblack}{RGB}{158,57,57}
\definecolor{mygold}{RGB}{150,112,35}

\title{Learning Reconstruction Errors as\\Spatiotemporal Evidence for\\AI-Generated Video Detection}

\author{Jiahua Jiang\thanks{Authors are listed in alphabetical order by surname.}\\
Institute of Mathematical Sciences\\
ShanghaiTech University\\
Shanghai, China\\
\texttt{jiangjh@shanghaitech.edu.cn}
\And
Jinglai Li\\
School of Mathematics\\
University of Birmingham\\
Birmingham B15 2TT, UK\\
\texttt{j.li.10@bham.ac.uk}
\AND
Jianru Zhang\\
School of Mathematics\\
University of Birmingham\\
Birmingham B15 2TT, UK\\
\texttt{jxz389@student.bham.ac.uk}
\And
Jin Zhu\\
School of Mathematics\\
University of Birmingham\\
Birmingham B15 2TT, UK\\
\texttt{j.zhu.7@bham.ac.uk}}

\iclrfinalcopy
\date{}
\hypersetup{hidelinks,pdfauthor={Jiahua Jiang, Jinglai Li, Jianru Zhang, Jin Zhu},pdftitle={Learning Reconstruction Errors as Spatiotemporal Evidence for AI-Generated Video Detection}}
\providecommand{\revise}[1]{#1}

\providecommand{\reviseblack}[1]{#1}

\begin{document}

\maketitle
\pagestyle{plain}
\thispagestyle{plain}

\begin{abstract}
The proliferation of AI-generated videos has raised growing concerns about misinformation, fraud, and potential societal risks, creating an urgent need for detection. Reconstruction-error-based methods have shown strong empirical performance for AI-generated image detection by exploiting discrepancies between an input and its reconstructed version. Their extension to video detection, however, faces two challenges: (i) a fixed reconstructor is not optimized for the detection objective and may yield weak discriminative signals; and (ii) an additional raw-video feature stream is often introduced to model temporal dependencies, increasing architectural overhead. To address these challenges, we develop a detector that learns and exploits reconstruction errors as a unified source of spatiotemporal evidence. First, we jointly optimize a reconstructor and the downstream classifier to produce reconstruction errors that are more informative for detecting AI-generated videos. {\color{black}Second, we show that consecutive reconstruction errors alone carry temporal evidence, and thus a 3D convolution over them captures spatiotemporal evidence without a raw-video stream.} Extensive experiments on over 30 existing AI-video generators demonstrate competitive performance against baselines.
\end{abstract}

\section{Introduction}

Nowadays, state-of-the-art video generation models, such as the closed-source models Sora and Veo2 and the open-source models CogVideoX and LTX-Video \citep{hacohen2024ltx, yang2024cogvideox}, can produce highly realistic videos when users provide text descriptions or input images. On the one hand, this makes video generation more convenient and thus enriches entertainment in ordinary life and the video/cinema business. On the other hand, misuse of video generation models can yield videos intended to spread misinformation and {facilitate} fraud, causing various risks to society and {businesses}. To avoid these potential risks, there is an important and increasingly urgent need to identify videos generated by AI.

Early AI-video detection methods \citep{qian2020thinking, frank2020leveraging, tan2024rethinking, yan2024df40} focus on DeepFake-style facial forgeries. More recently, state-of-the-art video generators can produce videos with diverse scenarios \citep{liu2025fakeparts, ma2026aigvdbench, ni2026genvidbench}, making detection more challenging. Various methods {for detecting such diverse AI-generated videos} have accordingly been developed \citep{chen2024demamba, liu2024turns, chen2026reconfuse, park2026vidguard}. Reconstruction-based methods are among the earliest techniques used to detect AI-generated videos. Specifically, a reconstruction-based method (a) reconstructs the video to be detected using a generative model, (b) computes the difference between the original and reconstructed videos, and (c) classifies the video as real or fake based on that difference. Reconstruction-based methods have demonstrated strong empirical performance in detecting AI-generated images \citep{cao2022recce, wang2023dire,luo2024lare2,chu2025fire,qi2026did}, and it is natural to extend this technique to AI-generated video detection.

However, reconstruction-based AI-video detection methods face two challenges. First, existing reconstruction-error-based video detectors often exploit the errors primarily as frame-wise spatial cues while {underusing} the temporal information in the errors \citep{liu2024turns, song2024learning, chen2026reconfuse}. As temporal information is important for detecting AI-generated videos \citep{chen2024demamba, zheng2025d3}, existing methods introduce additional features from the raw video to supplement the information provided by reconstruction errors. Second, when real and generated videos produce similar reconstruction-error patterns, the errors provide only a weak discriminative signal, limiting their predictive power. These observations raise a central question:
\begin{center}
\textit{Can we produce {more discriminative} reconstruction errors and exploit them as a unified source of spatial and temporal evidence for AI-generated video detection?}
\end{center}

To answer this question, we develop a detector that learns and exploits reconstruction errors within a single stream. First, we jointly optimize a reconstructor and the downstream classifier to produce reconstruction errors that are more informative for detecting AI-generated videos. {\color{black}Second, we show that consecutive reconstruction
errors contain temporal detection cues. We therefore apply3D convolution directly to the error sequence to jointly
capture spatial and temporal evidence, without an additional
raw-video branch.} Empirically, we test the proposal on several benchmark datasets covering more than 30 existing AI-video generators. {The evaluated generators include both text-to-video and image-to-video models.} Across these experiments, even when using only reconstruction errors, we find that the proposal typically achieves the highest {detection performance} when compared with more than 10 baselines.

\section{Related Work}
\revise{This section mainly reviews existing literature on detecting AI-generated videos, as well as the reconstruction-based detection approaches that are most relevant to our method. Additional literature on approaches for generating AI videos is reviewed in Appendix~\ref{sec:addition-related-work}.}

\textbf{AI-generated Video Detection{.}}
{Existing} AI-generated video detectors are mainly {training-based methods} that {leverage datasets containing real and AI-generated videos} to train {classifiers}~\citep{corvi2026seeing, li2026preserving}. {Training-based methods can be categorized} into several subtypes. Inconsistency-based detection methods typically use {pretrained CLIP-type models}~\citep{radford2021learning,ni2022expanding} to extract features, {which are then used} to train a binary classifier to identify {discrepancies} between {AI-generated and real videos in the spatial domain, the temporal domain, or both}~\citep{song2024learning, chen2024demamba, ma2024decof, xiang2026semadet, wei2026specsem}. Furthermore, {methods in this category may also incorporate reconstruction errors, the differences between observed videos and their reconstructed counterparts, as additional features} for training the classifier~\citep{liu2024turns, chen2026reconfuse, du2026g2vd}. {Reasoning-based methods}~\citep{wen2025busterx, li2026skyra, park2026vidguard, tan2026videoveritas} {exploit high-capacity pretrained} multimodal large language {models and use post-training techniques,} such as {supervised fine-tuning} and reinforcement learning{,} to detect {forgeries} and provide {explanations}. {Physics-driven methods}~\citep{interno2025straightening, he2026motionphys, zhang2025physics} {identify AI-generated videos by detecting deviations from natural motion}. Finally, although most methods are training-based, {training-free methods exploit second-order features, spatiotemporal likelihoods, or temporal over-coherence for detection~\citep{zheng2025d3, benhayun2026stall, brokman2026training}.}

\textbf{Reconstruction-Based {AI-Generated} Image/Video Detection}.
\underline{Di}ffusion \underline{Re}construction (DIRE){-based} detection exploits discrepancies between an input and its AI-reconstructed counterpart as forensic evidence. {\citet{wang2023dire}} found {that real and synthetic images exhibit} systematically different DIRE, and thus, {DIRE-based detection performs well} in AI-generated image detection. Motivated by the success of DIRE, \citet{ricker2024aeroblade, luo2024lare2} proposed fast {variants} of DIRE to make it more {practical to use}. Furthermore, new methods enrich {the} reconstruction signal to enhance the detection power through diffusion inversion \citep{cazenavette2024fakeinversion}, semantically guided image reconstruction {\citep{jin2026dysy,yu2024semgir}}, frequency decomposition \citep{chu2025fire}, structural edits in reconstruction error \citep{jiang2026eires}, the second-order changes in reconstruction error {\citep{qi2026did}}, and {correction of} instance-specific biases in reconstruction error {\citep{choi2026debiased}}. {Reconstruction-based methods are currently also among the} important {baselines} in video {detection} \citep{liu2024turns, song2024learning, chen2026reconfuse,du2026g2vd}.

\section{Preliminaries}\label{sec:preliminary}
In this section, we would review two AI video detection approaches that motivate our method: \underline{\textit{reconstruction-based method}} and \underline{\textit{the second-order difference method}}. Throughout the paper, we denote a video as $x \in \mathbb{R}^{T\times C \times H \times W}$ where $T$ is the number of frames, $C$ is the number of channels, and $H$ and $W$ are the frame height and width, respectively. The $t$-th frame is denoted by
$x_t \in \mathbb{R}^{C\times H\times W}$, for $t=1,\ldots,T$.

\begin{figure}[htbp]
\centering
\begin{minipage}[l]{0.3\linewidth}
\begin{flushleft}
\textbf{(a)}
\end{flushleft}
\vspace{-6.3pt}
\includegraphics[width=\linewidth]{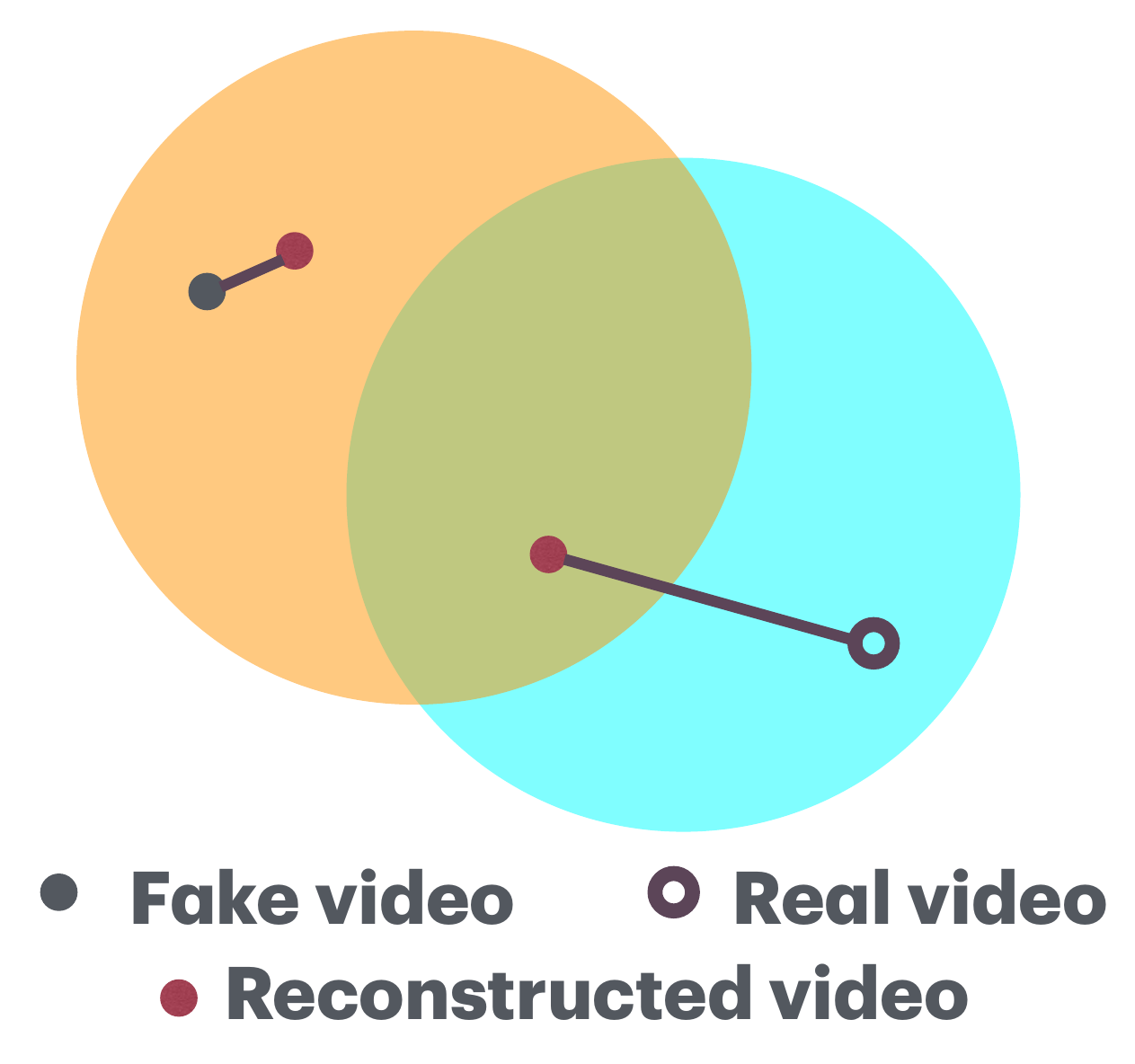}
\end{minipage}
\begin{minipage}[r]{0.69\linewidth}
\begin{flushleft}
\textbf{(b)}
\end{flushleft}
\vspace{-6pt}
\includegraphics[width=\linewidth]{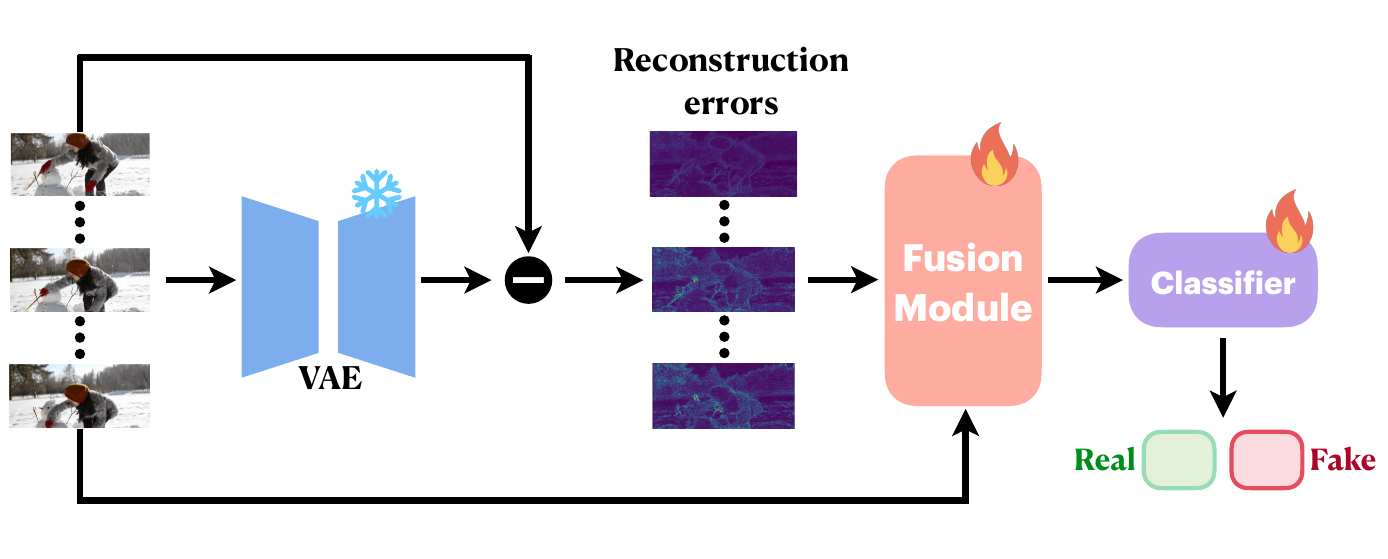}
\end{minipage}
\vspace*{-8pt}
\caption{Panel (a). Graphical illustration of the rationale behind existing reconstruction-error-based methods. Panel (b). A high-level workflow of existing reconstruction-error-based methods.}\label{fig:review-dire}
\end{figure}

{\color{black}\textbf{Reconstruction-based method.}
A key component is the reconstructor $\mathcal{R}(\cdot)$, which maps
an observed video $x$ to its reconstruction. Then, the detectors exploit the reconstruction error
\begin{align*}
x-\mathcal{R}(x) \in \mathbb{R}^{T \times C \times H \times W}
\end{align*}
as forensic evidence for detecting AI-generated videos. Typically, $\mathcal{R}$ is implemented with an variational auto-encoder (VAE) \citep{song2024learning, chen2026reconfuse}. The underlying rationale is that AI-generated videos is associted with a larger reconstruction errors magnitude (see Figure~\ref{fig:review-dire}(a) for an illustration).} The resulting error forms a sequence of frame-wise discrepancies. However, existing approaches often use it primarily as a spatial cue and introduce an additional raw-video branch to model temporal dependencies \citep{liu2024turns, song2024learning, chen2026reconfuse}. Specifically, the raw video and its reconstruction error are concatenated and fused via a dedicated fusion module, as shown in Figure~\ref{fig:review-dire}(b). The fused information is then used to predict whether the video is real or fake.

\begin{wrapfigure}{r}{0.5\textwidth}
\vspace*{-16pt}
\centering
\includegraphics[width=1.0\linewidth]{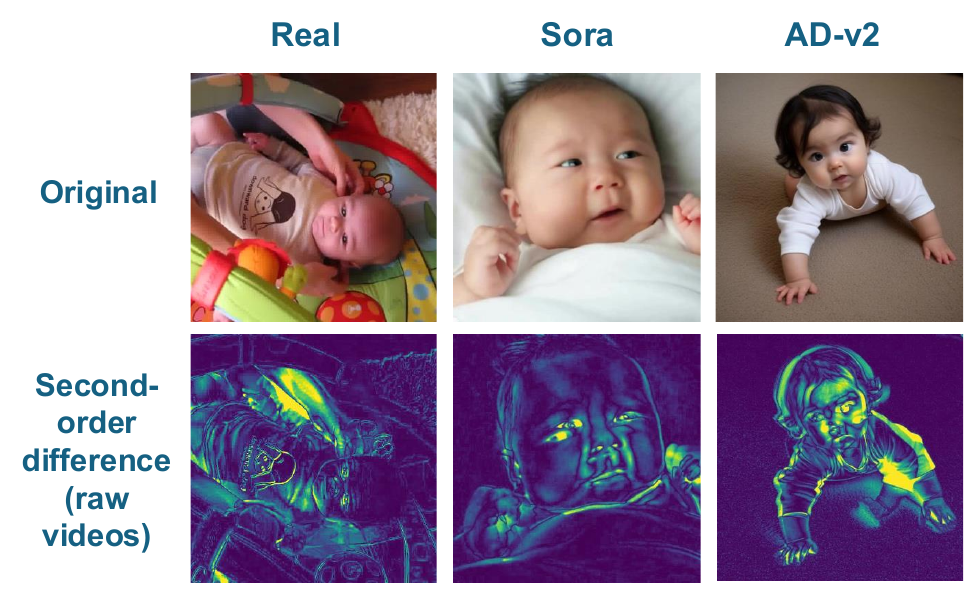}   
\vspace*{-16pt} 
\caption{Comparison of $\Delta_t^2 x_t$ in real videos (left panels) and fake videos generated by Sora (middle panels) and AnimateDiff-v2 (right panels). }\label{fig:2nd-diff}
\vspace*{-13pt} 
\end{wrapfigure}
\textbf{Second-order difference method}. As noted above, AI-generated and natural videos can exhibit different temporal behaviors, and existing studies have shown that such differences are informative for AI-generated video detection. {\color{black}Specifically, \citet{zheng2025d3} propose that the second-order temporal features, i.e., given a sequence of video frames $x_1,\ldots,x_T$, they define the approximation of the acceleration at time $t$ as the second-order difference between raw video frames:
\begin{align*}
\Delta_t^2 x_t &= (x_t - x_{t-1}) - (x_{t-1} - x_{t-2}) 
\\
&= x_t-2x_{t-1}+x_{t-2}.
\end{align*}
Figure~\ref{fig:2nd-diff} provides an illustration of this signal for discriminating real and fake videos.} For AI-generated videos, the second-order differences concentrate on the moving subject, while for real videos, the differences appear in both the moving subject and the background. This forms the practical principle for second-order difference-based detection.

\section{Methodology}\label{sec:method}

\subsection{Motivation}
\begin{wrapfigure}{r}{0.5\textwidth}
\vspace*{-32pt}
\centering
\includegraphics[width=1.0\linewidth]{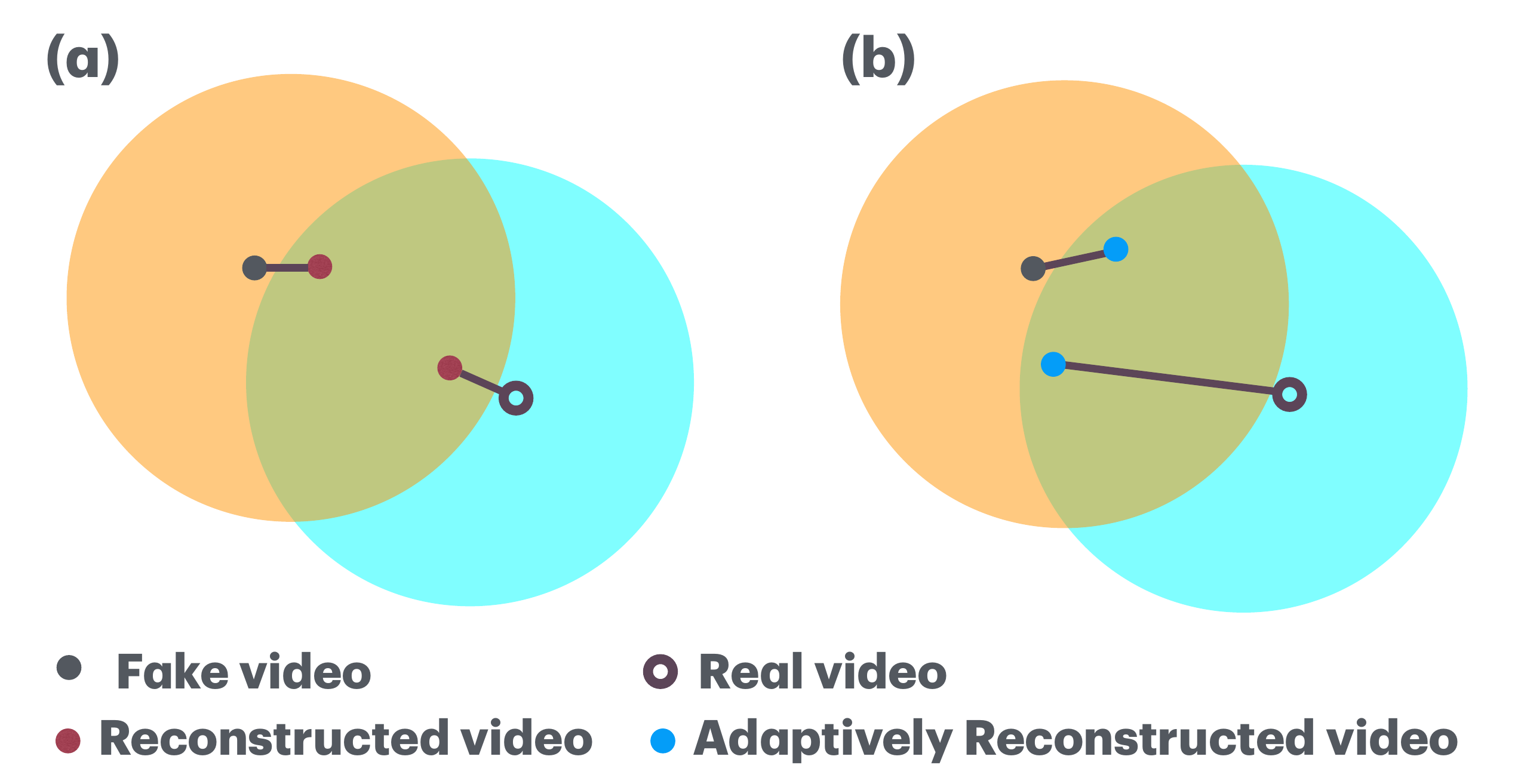}
\vspace*{-16pt}
\caption{Panel (a): the reconstruction {errors obtained with} a fixed reconstructor.
Panel (b): the reconstruction errors {obtained with} an adaptive reconstructor.
Adaptation can make reconstruction-error patterns more distinguishable.}
\label{fig:adadire}
\vspace*{-13pt}
\end{wrapfigure}

This paper asks whether we can adaptively learn and exploit reconstruction errors as a unified source of spatial and temporal evidence for AI-generated video detection. We begin by discussing two limitations of existing implementations.
\begin{itemize}[leftmargin=*]
\item \textit{Discriminativeness of reconstruction errors}. Existing reconstruction-based detectors typically treat $\mathcal{R}$ as fixed \citep{wang2023dire,ricker2024aeroblade,chen2026reconfuse,du2026g2vd}. However, as the fixed reconstructor is not optimized for the detection objective, the reconstruction errors may not be sufficiently distinguishable, particularly when real videos closely resemble AI-generated videos and vice versa (see Figure~\ref{fig:adadire}(a)). In such cases, the reconstruction errors provide only a weak discriminative signal to the subsequent classifier, limiting its predictive power.
\item \textit{Neglect of temporal signals in reconstruction errors}. Reconstruction errors form a temporal sequence across video frames, but existing methods often use them primarily as frame-wise spatial cues rather than directly modeling their temporal structure. To compensate for this underuse, these methods introduce an additional raw-video feature-extraction branch and a fusion module to incorporate temporal information, thereby increasing architectural and inference overhead.
\end{itemize}
\begin{figure}[H]
\centering
\vspace{-8pt}
\includegraphics[width=0.9\linewidth]{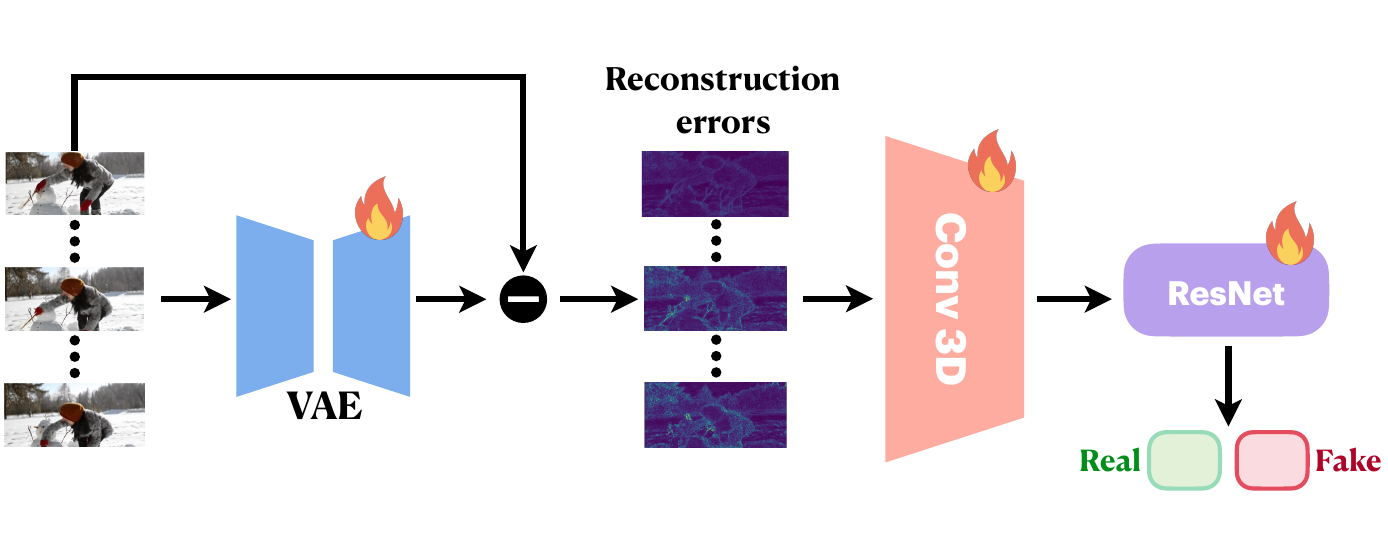}
\vspace{-18pt}
\caption{Overview of the proposed detection framework. An adaptive variational autoencoder produces reconstruction errors for detection. \textcolor{black}{A 3D convolution over consecutive errors extracts temporal evidence and integrates it with spatial evidence.}}
\vspace{-8pt}
\label{fig:method}
\end{figure}
Together, these limitations concern how reconstruction errors are produced and used. Our method addresses both through (i) an adaptive reconstructor that produces more discriminative reconstruction errors and (ii) {\color{black}a 3D convolution that extracts temporal evidence from consecutive reconstruction errors and integrates it with spatial evidence}. Figure~\ref{fig:method} gives an overview of the {proposed method}, with the details of (i) and (ii) presented in Sections~\ref{sec:adadire} and~\ref{sec:high-order-diff}, respectively. \revise{The implementation details are provided in Appendix~\ref{sec:implement-details}.}

\subsection{Adaptive Reconstruction Errors}\label{sec:adadire}

We propose learning adaptive reconstruction errors that {provide} a more discriminative signal for detecting AI-generated videos. Specifically, suppose that we have access to a dataset of real videos $\mathcal{D}_r$ and a dataset of AI-generated videos $\mathcal{D}_f$, both of which are readily available in practice. For instance, $\mathcal{D}_r$ can be a pre-collected video dataset, while $\mathcal{D}_f$ can be constructed by prompting a text-to-video model to generate videos. {We} parameterize the reconstructor as $\mathcal{R}_{\psi}(\cdot)$, where $\psi$ comprises all learnable parameters in the reconstructor. For an input video $x$, the adaptive reconstructor produces the learned reconstruction error $x-\mathcal{R}_{\psi}(x) \in\mathbb{R}^{T\times C\times H\times W}$. We then feed the error into a classifier $\mathcal{C}_{\phi}$, where $\phi$ comprises all parameters in the classifier. We {interpret $\mathcal{C}_{\phi}(x-\mathcal{R}_{\psi}(x))\in(0,1)$} as the predicted probability that $x$ is AI-generated. We jointly learn $\phi$ and $\psi$ by minimizing the binary cross-entropy loss
\begin{equation}\label{eq:loss}
\begin{split}
\mathcal{L}(\phi,\psi)
&=-\sum_{x\in\mathcal{D}_f}
\log \mathcal{C}_{\phi}(x-\mathcal{R}_{\psi}(x))
-\sum_{x\in\mathcal{D}_r}
\log\left[1-\mathcal{C}_{\phi}(x-\mathcal{R}_{\psi}(x))\right].
\end{split}
\end{equation}

\begin{wrapfigure}{l}{0.55\textwidth}
\centering
\vspace{-16pt}
\includegraphics[width=1.0\linewidth]{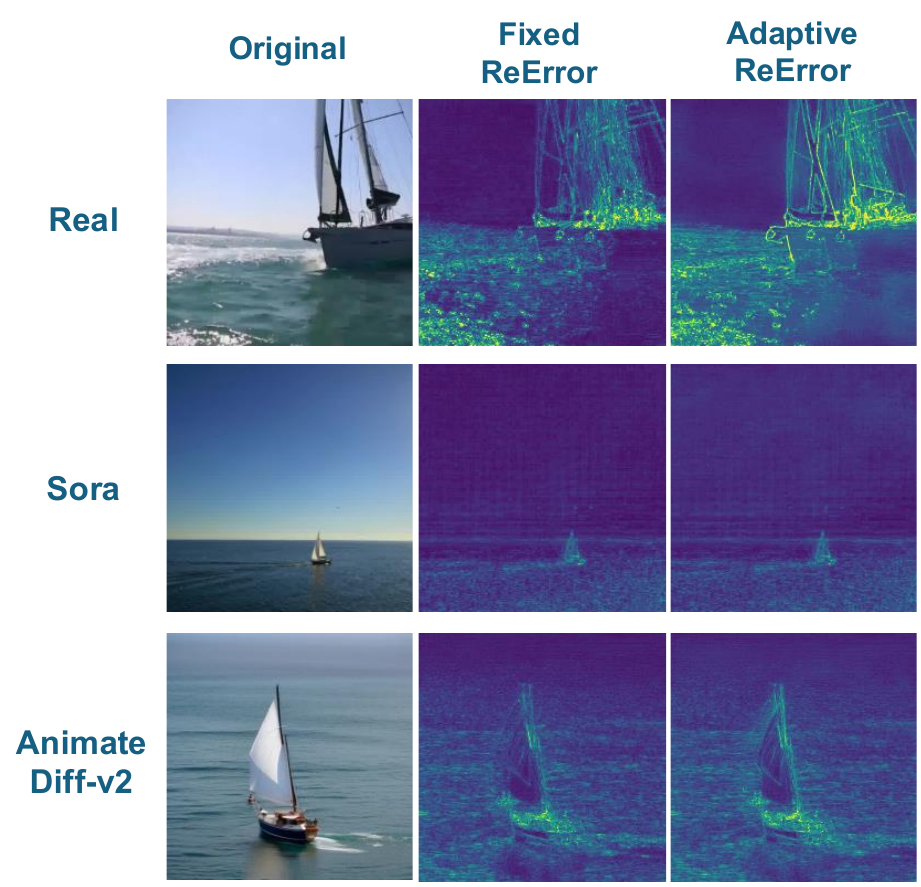}
\vspace{-12pt}
\caption{Graphical comparison of the reconstruction errors returned by a fixed reconstructor (middle column) and an adaptive reconstructor (right column) for the original videos (left column). The first row shows a real video, while the second and third rows show videos from Sora and AnimateDiff-v2, respectively.}
\label{fig:motivate-lr}
\end{wrapfigure}
To instantiate the learnable reconstructor in practice,
we initialize $\mathcal{R}_{\psi}$ from a pretrained VAE and adapt it via low-rank adaptation
\citep[LoRA;][]{hu2022lora}, freezing the pretrained weights and optimizing only the introduced LoRA parameters.
The pretrained initialization provides a reconstruction-oriented
{starting point}, while LoRA keeps the overall number
of {trainable} parameters small.
Since {$\mathcal{C}_{\phi}(x-\mathcal{R}_{\psi}(x))$}
maps an input video directly to a classification probability
and both $\phi$ and $\psi$ are optimized using the detection loss,
the proposed detector can be viewed as an end-to-end error-learning model.
Algebraically, {$x-\mathcal{R}_{\psi}(x)
=x+[-\mathcal{R}_{\psi}(x)]$ has} an identity-plus-learned-branch form
analogous to that used in residual learning~\citep{he2016deep}.
Such additive skip connections are also used in Transformer and YOLOv3 architectures~\citep{vaswani2017attention, redmon2018yolov3}.

Figure~\ref{fig:motivate-lr} provides an illustrative example of the resulting change in reconstruction errors. {For the frames shown}, adaptive reconstruction changes the error pattern more visibly for the real video than for the two generated videos. In this example, the resulting {error patterns} are therefore easier to distinguish. The figure provides qualitative evidence for this intended mechanism. The detection loss encourages the adaptive reconstructor to produce reconstruction-error patterns that better distinguish real and
generated videos.

\subsection{\reviseblack{Spatiotemporal Modeling of Reconstruction Errors}}
\label{sec:high-order-diff}

\begin{wrapfigure}{r}{0.5\textwidth}
\vspace*{-18pt}
\centering
\includegraphics[width=1.0\linewidth]{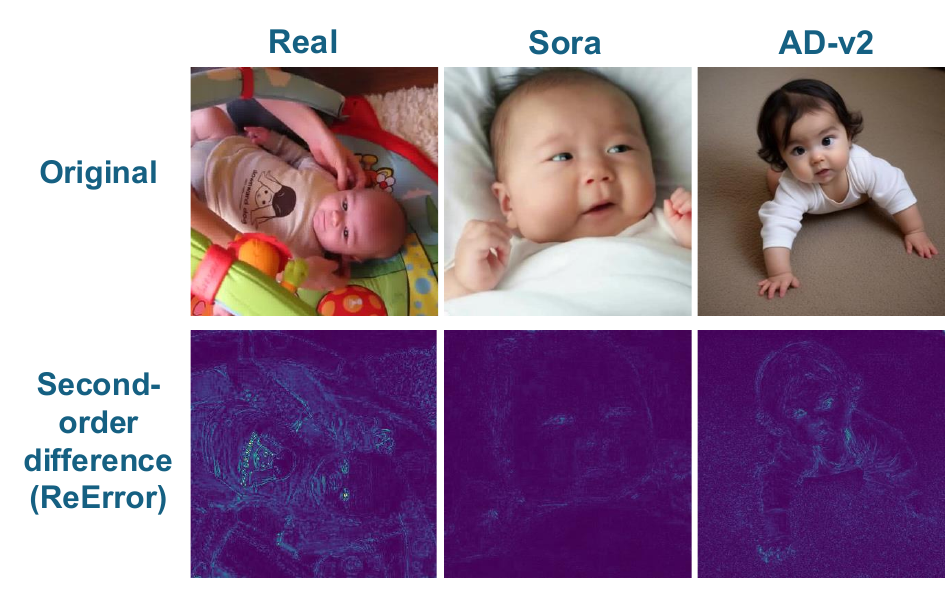}
\vspace*{-18pt}
\caption{Comparison of $\Delta_t^2 x_t-\Delta_t^2\widetilde{x}_t$ in real videos (left column) and fake videos generated by Sora (middle column) and AnimateDiff-v2 (right column).}
\label{fig:hessian-diff}
\vspace{-12pt}
\end{wrapfigure}
\reviseblack{
Our detector applies 3D convolution directly to consecutive reconstruction errors. We motivate this design through an analysis of temporal cues in the error sequence, explain how learnable kernels flexibly integrate these temporal cues with spatial evidence, and validate the design through ablation studies.

To examine the temporal detection cues in the reconstruction error sequence, we consider the second-order temporal difference method discussed in Section~\ref{sec:preliminary}. Specifically, let $r_t=x_t-\widetilde{x}_t$ denote the reconstruction error at frame $t$ where $\widetilde{x}_t$ is the reconstruction of $x_t$, we consider the difference between the second-order temporal differences of a video and its reconstruction:
\begin{equation}
\Delta_t^2 x_t-\Delta_t^2\tilde{x}_t
=
r_t-2r_{t-1}+r_{t-2}.
\end{equation}
As illustrated in Figure~\ref{fig:hessian-diff}, this signal is spread across the real-video frame but concentrated around moving objects in the generated examples. This contrast suggests that reconstruction errors contain useful temporal detection cues beyond framewise spatial patterns.

The second-order difference is one particular way to combine
consecutive maps. More generally, to learn more discriminative
temporal combinations, we apply a 3D convolution to consecutive
reconstruction errors $\{r_t\}$, yielding
\begin{equation}
W_1 * r_t
+ W_2 * r_{t-1}
+ W_3 * r_{t-2},
\label{eq:3d-conv-approach}
\end{equation}
where each $W_i$ is a learnable 2D spatial kernel and $*$ denotes 2D convolution summed over input channels. Setting the three kernel slices to $1$, $-2$, and $1$ times the identity kernel yields $\Delta_t^2 x_t-\Delta_t^2\widetilde{x}_t$. Rather than using this fixed second-order signal, our classifier learns $W_1, W_2, W_3$ under the detection objective for flexible spatiotemporal modeling. Each kernel slice extracts local spatial patterns, while their learnable weighted combination integrates temporal information across consecutive frames. This allows spatial and temporal evidence to be learned within a single stream, without an additional raw-video branch or cross-branch fusion.

Finally, the ablations in Table~\ref{tab:ablation} show that our 3D convolutional classifier achieves higher average performance than framewise classification (variant (b)) and the alternatives using explicit second-order difference signals (variants (c) and (d)), supporting its effectiveness in learning spatiotemporal evidence directly from reconstruction errors.
}

\section{Experiments}\label{sec:experiments}

\subsection{Experimental Setup}

\revise{\textbf{Implementation: high-level description.} We adapt Wan2.1-VAE with rank-$32$ LoRA in eight Conv3D layers within the encoder and decoder middle blocks, keeping the pretrained weights frozen. The reconstruction errors are fed to a randomly initialized 3D ResNet-18-style classifier. We use the kernel with the same shape as existing 3D ResNet\footnote{\url{https://docs.pytorch.org/vision/stable/_modules/torchvision/models/video/resnet.html}}. The LoRA adapters and classifier are jointly trained with binary cross-entropy loss (see Equation~(\ref{eq:loss})). Appendix~\ref{sec:implement-details} provide details such as reconstructor, classifier, and optimization.}

\textbf{Baselines.} We compare our method with 11 representative AI-generated video detectors, most of which are training-based detectors, including ReConFuse \citep{chen2026reconfuse}, two DIVID variants \citep{liu2024turns} (DIVID-CNN and DIVID-LSTM), G2VD-CLIP \citep{du2026g2vd}, DeMamba \citep{chen2024demamba}, Ivy-xDetector \citep{ivy2026jiang}, DeepTraceReward \citep{fu2025learning}, and MM-Det \citep{song2024learning}. We also compare it with three training-free detectors: D3 \citep{zheng2025d3}, STALL \citep{benhayun2026stall}, and Over-Coherence \citep{brokman2026training}.

\textbf{Training setup.} We train the proposed method on 139,002 class-balanced videos: 69,501 real videos from Kinetics-400 \citep{kay2017kinetics} and 69,501 synthetic videos generated in the VBench dataset \citep{huang2023vbench,huang2025vbench}. The checkpoint is selected using a hold-out validation set that includes fake videos from the MovieGen dataset \citep{polyak2024movie} and real videos from the Panda-70M dataset \citep{chen2024panda70m}. \revise{Baselines without available checkpoints are retrained on our training corpus, using method-specific training budgets detailed in Appendix~\ref{sec:experiments-details}; all methods use the same evaluation sets.}

\textbf{Evaluation.} We evaluate all the baselines on datasets that were not used during training or validation, including (i) a dataset mixed with fake videos from the EvalCrafter dataset \citep{liu2024evalcrafter} and real videos from the MSR-VTT dataset \citep{xu2016msr}, (ii) the GenVideo dataset that has both real and fake videos \citep{chen2024demamba}, (iii) a dataset mixed with fake videos from the FakeParts dataset \citep{liu2025fakeparts} and real videos from the TenKReal and ROVI datasets \citep{wu2024towards}, (iv) a dataset mixed with fake videos from VideoPhy \citep{bansal2025videophy} and real videos from MSR-VTT and \revise{(v) a dataset with highly realistic faked videos (HRFVideo) from recent generators such as Seedance \citep{seedance2026seedance}, denoted as the HRFVideo dataset.} These datasets collectively over 30 synthetic-video subsets spanning text-to-video and image-to-video generation. We report video-level AUC, and average precision (AP) that treats generated videos as the positive class. 

\subsection{Comparison to existing methods}

First, from the results on the EvalCrafter dataset (see Table~\ref{tab:evalcrafter-auc}), we can see that our method has the highest AUC for 10 of the 14 generators and still ranks second for {3 generators and third for ModelScope}. This leads to the best overall AUC on the EvalCrafter dataset, which exceeds 0.99 on average. Notably, it uniformly outperforms the other reconstruction-based baselines, such as ReConFuse, MM-Det, and two DIVID variants. These results demonstrate the effectiveness of our proposal on leveraging the reconstruction errors. In terms of the AP results presented in Table~\ref{tab:evalcrafter-ap}, our method still delivers the leading performance. For 12 of the 14 generators, our method has the highest AP, and for the remaining generators, it has the second-best AP. The empirical superiority of the proposal extends to the GenVideo and VideoPhy datasets. It typically has the best performance across video generators in the dataset (as presented in Table~\ref{tab:genvideo} and Table~\ref{tab:videophy-ap}), demonstrating the empirical advantages of the proposal. \revise{Finally, we compare the methods on two more challenging datasets, FakeParts and HRFVideo. In FakeParts, some AI-generated videos are produced by fine-grained manipulations, such as style changes and inpainting. The results are reported in Table~\ref{tab:fakeparts} and Table~\ref{tab:hrfvideo_reconstruction}. On the two datasets, our method shows overall competitive performance and achieves the highest average AUC and AP.}

\begin{table}[t]
\centering
\caption{AUC scores across generators in the EvalCrafter datasets; the best method is highlighted in \colorbox{cyan!24}{cyan}, the second-best method in \colorbox{orange!24}{orange}. Each column represents one video generator and each row is one method. \revise{Ours reports mean $\pm$ standard deviation over three independent runs.}}\label{tab:evalcrafter-auc}
\begin{adjustbox}{max width=\textwidth}
\begin{tabular}{lccccccc}
\toprule
& {floor33} & {Gen-2} & {Gen-2 Dec.} & {hotshot} & {LaVIE base} & {LaVIE interp.} & {ModelScope} \\
\midrule
D3 & 65.00 & 86.49 & 88.89 & 85.43 & 73.02 & 73.97 & 61.74 \\
STALL & 79.43 & 86.70 & 89.55 & 73.33 & 84.66 & 81.88 & 73.35 \\
Over-Coherence & 47.55 & 32.07 & 49.01 & 70.92 & 45.44 & 33.90 & 42.55 \\
ReConFuse & 98.61 & 99.05 & 99.26 & \cellcolor{orange!24} 97.08 & 98.09 & 97.16 & 96.78 \\
DIVID-CNN & 95.94 & 97.44 & 98.52 & 83.43 & 96.86 & 93.80 & 92.83 \\
DIVID-LSTM & 95.07 & 96.70 & 97.72 & 70.26 & 94.48 & 91.14 & 87.28 \\
G2VD-CLIP & \cellcolor{orange!24} 99.57 & 99.45 & 99.47 & 93.41 & \cellcolor{orange!24} 98.78 & 96.29 & 97.57 \\
Ivy-xDetector & 94.17 & 95.05 & 98.16 & 93.30 & 90.91 & 81.67 & 90.72 \\
DeepTraceReward & 99.03 & \cellcolor{cyan!24} 99.82 & \cellcolor{cyan!24} 99.92 & 82.65 & 84.77 & 93.69 & 96.43 \\
MM-Det & 55.09 & 58.21 & 65.23 & 59.82 & 50.95 & 56.74 & 55.11 \\
DeMamba & 99.36 & 99.61 & 99.36 & 96.88 & 98.53 & \cellcolor{orange!24} 98.28 & \cellcolor{orange!24} 97.79 \\
\revise{Ours} & \cellcolor{cyan!24} \revise{$99.95\pm0.02$} & \cellcolor{orange!24} \revise{$99.77\pm0.07$} & \cellcolor{orange!24} \revise{$99.89\pm0.08$} & \cellcolor{cyan!24} \revise{$99.96\pm0.05$} & \cellcolor{cyan!24} \revise{$99.98\pm0.04$} & \cellcolor{cyan!24} \revise{$99.81\pm0.19$} & \cellcolor{cyan!24} \revise{$99.00\pm1.66$} \\
\midrule
& {Moonvalley} & {Pika} & {Pika Dec.} & {show\_1} & {VideoCrafter2} & {VideoCrafter1} & {zeroscope} \\
\midrule
D3 & 85.10 & 82.46 & 85.32 & 76.24 & 86.49 & 80.91 & 56.10 \\
STALL & 78.63 & 83.48 & 87.60 & 78.85 & 85.40 & 82.77 & 66.04 \\
Over-Coherence & 43.24 & 32.51 & 38.83 & 28.59 & 55.06 & 57.32 & 42.22 \\
ReConFuse & 99.65 & 98.85 & 99.00 & 97.31 & 99.33 & 98.93 & 92.87 \\
DIVID-CNN & 99.24 & 97.32 & 97.24 & 88.50 & 99.14 & 97.45 & 79.19 \\
DIVID-LSTM & 99.17 & 96.72 & 96.94 & 90.77 & 98.47 & 96.38 & 71.48 \\
G2VD-CLIP & 99.85 & 99.41 & 99.40 & 95.52 & 99.86 & 99.24 & 93.63 \\
Ivy-xDetector & 98.70 & 94.73 & 94.18 & 85.38 & 98.67 & 96.45 & 87.20 \\
DeepTraceReward & 99.54 & \cellcolor{orange!24} 99.86 & \cellcolor{orange!24} 99.85 & 86.91 & \cellcolor{cyan!24} 99.99 & \cellcolor{orange!24} 99.25 & 91.82 \\
MM-Det & 62.98 & 53.75 & 51.45 & 51.94 & 50.87 & 52.67 & 50.22 \\
DeMamba & \cellcolor{orange!24} 99.86 & 99.78 & 98.72 & \cellcolor{orange!24} 98.83 & 99.69 & 99.12 & \cellcolor{orange!24} 97.15 \\
\revise{Ours} & \cellcolor{cyan!24} \revise{$99.89\pm0.10$} & \cellcolor{cyan!24} \revise{$99.89\pm0.07$} & \cellcolor{cyan!24} \revise{$99.97\pm0.02$} & \cellcolor{cyan!24} \revise{$99.93\pm0.07$} & \cellcolor{orange!24} \revise{$99.96\pm0.04$} & \cellcolor{cyan!24} \revise{$99.90\pm0.09$} & \cellcolor{cyan!24} \revise{$99.96\pm0.03$} \\
\bottomrule
\end{tabular}
\end{adjustbox}
\end{table}

\begin{table}[t]
\centering
\caption{AP scores across generators in the EvalCrafter datasets.}
\begin{adjustbox}{width=\textwidth}
\begin{tabular}{lccccccc}
\toprule
Model & {floor33} & {Gen-2} & {Gen-2 Dec.} & {hotshot} & {LaVIE base} & {LaVIE interp.} & {ModelScope} \\
\midrule
D3 & 36.89 & 55.37 & 62.85 & 54.81 & 46.07 & 45.56 & 31.14 \\
STALL & 58.78 & 66.77 & 77.06 & 49.37 & 68.89 & 64.03 & 48.71 \\
Over-Coherence & 23.18 & 17.89 & 23.55 & 38.18 & 23.16 & 19.09 & 22.36 \\
ReConFuse & 96.81 & 97.79 & 98.29 & \cellcolor{orange!24} 93.02 & 95.61 & 93.49 & 92.19 \\
DIVID-CNN & 93.81 & 95.83 & 97.52 & 74.03 & 94.27 & 88.87 & 86.30 \\
DIVID-LSTM & 91.74 & 92.68 & 95.42 & 51.77 & 89.50 & 82.29 & 75.21 \\
G2VD-CLIP & \cellcolor{orange!24} 98.59 & 98.24 & 98.26 & 85.11 & \cellcolor{orange!24} 97.00 & 91.70 & 94.04 \\
Ivy-xDetector & 82.29 & 83.11 & 94.97 & 81.48 & 74.71 & 57.59 & 73.29 \\
DeepTraceReward & 97.94 & \cellcolor{cyan!24} 99.58 & \cellcolor{orange!24} 99.79 & 63.67 & 68.28 & 86.56 & 93.84 \\
MM-Det & 30.66 & 34.38 & 45.47 & 37.31 & 26.35 & 32.96 & 31.35 \\
DeMamba & 98.39 & 98.92 & 98.19 & 91.64 & 95.93 & \cellcolor{orange!24} 95.18 & \cellcolor{orange!24} 94.47 \\
\revise{Ours} & \cellcolor{cyan!24} \revise{$99.90\pm0.03$} & \cellcolor{orange!24} \revise{$99.76\pm0.36$} & \cellcolor{cyan!24} \revise{$99.89\pm0.09$} & \cellcolor{cyan!24} \revise{$99.87\pm0.11$} & \cellcolor{cyan!24} \revise{$99.98\pm0.02$} & \cellcolor{cyan!24} \revise{$99.64\pm0.47$} & \cellcolor{cyan!24} \revise{$98.42\pm2.63$} \\
\midrule
Model & {Moonvalley} & {Pika} & {Pika Dec.} & {show\_1} & {VideoCrafter2} & {VideoCrafter1} & {zeroscope} \\
\midrule
D3 & 54.47 & 55.74 & 60.71 & 48.55 & 60.14 & 57.84 & 24.22 \\
STALL & 52.58 & 66.89 & 71.73 & 57.22 & 66.58 & 66.34 & 33.05 \\
Over-Coherence & 19.65 & 18.32 & 19.93 & 17.74 & 26.26 & 28.11 & 21.07 \\
ReConFuse & 98.95 & 96.99 & 97.48 & 93.62 & 98.68 & 97.30 & 81.65 \\
DIVID-CNN & 98.77 & 95.33 & 95.75 & 80.77 & 98.60 & 95.72 & 58.56 \\
DIVID-LSTM & 98.20 & 93.30 & 94.76 & 82.01 & 97.44 & 94.22 & 49.08 \\
G2VD-CLIP & 99.37 & 98.09 & 98.09 & 88.90 & 99.51 & 97.99 & 84.33 \\
Ivy-xDetector & 95.02 & 82.70 & 79.08 & 58.48 & 95.59 & 89.48 & 58.39 \\
DeepTraceReward & 98.82 & \cellcolor{orange!24} 99.61 & \cellcolor{orange!24} 99.66 & 73.65 & \cellcolor{cyan!24} 99.99 & \cellcolor{orange!24} 98.21 & 84.31 \\
MM-Det & 40.81 & 29.09 & 26.88 & 27.01 & 26.28 & 28.06 & 23.10 \\
DeMamba & \cellcolor{orange!24} 99.59 & 99.35 & 96.65 & \cellcolor{orange!24} 96.53 & 99.26 & 97.58 & \cellcolor{orange!24} 89.91 \\
\revise{Ours} & \cellcolor{cyan!24} \revise{$99.95\pm0.05$} & \cellcolor{cyan!24} \revise{$99.83\pm0.14$} & \cellcolor{cyan!24} \revise{$99.88\pm0.02$} & \cellcolor{cyan!24} \revise{$99.86\pm0.08$} & \cellcolor{orange!24} \revise{$99.90\pm0.10$} & \cellcolor{cyan!24} \revise{$99.93\pm0.06$} & \cellcolor{cyan!24} \revise{$99.89\pm0.11$} \\
\bottomrule
\end{tabular}
\end{adjustbox}
\label{tab:evalcrafter-ap}
\end{table}

\begin{table}[t]
\centering
\caption{AUC (upper part) and AP (lower part) on the GenVideo dataset. Avg. denotes the mean across the generators.}
\small
\begin{adjustbox}{max width=\textwidth}
\begin{tabular}{lcccccccccc}
\toprule
 & {Crafter} & {Gen-2} & {LaVIE} & {ModelScope} & {Moonvalley} & {MorphStudio} & {Show-1} & {Sora} & {Wildscrape} & Avg. \\
\midrule
D3 & 98.24 & 99.13 & 97.24 & 95.23 & 99.00 & 97.58 & 98.36 & 97.12 & \cellcolor{cyan!24} 94.66 & \cellcolor{orange!24} 97.40 \\
STALL & 85.98 & 90.25 & 79.70 & 72.40 & 81.14 & 79.22 & 77.97 & 76.00 & 68.44 & 79.01 \\
Over-Coherence & 27.12 & 10.73 & 29.60 & 35.81 & 17.60 & 25.62 & 20.17 & 23.44 & 22.83 & 23.66 \\
ReConFuse & 98.64 & 99.00 & 96.71 & 96.65 & 99.40 & 98.03 & 97.55 & 94.41 & 89.95 & 96.70 \\
DIVID-CNN & 95.17 & 94.96 & 87.76 & 86.47 & 98.39 & 93.48 & 78.90 & 89.35 & 70.08 & 88.28 \\
DIVID-LSTM & 91.31 & 88.90 & 82.45 & 75.61 & 95.81 & 88.42 & 77.35 & 74.04 & 61.30 & 81.69 \\
G2VD-CLIP & 99.19 & 99.26 & 93.81 & 96.43 & 99.82 & \cellcolor{orange!24} 99.45 & 89.49 & 96.41 & 82.90 & 95.20 \\
Ivy-xDetector & 97.75 & 96.32 & 86.67 & 90.76 & 98.76 & 94.64 & 85.59 & 92.84 & 91.69 & 92.78 \\
DeepTraceReward & \cellcolor{orange!24} 99.55 & \cellcolor{orange!24} 99.97 & 88.19 & 96.82 & 99.49 & 98.80 & 87.56 & \cellcolor{orange!24} 99.17 & 89.28 & 95.43 \\
MM-Det & 51.73 & 62.41 & 53.80 & 54.29 & 63.35 & 55.89 & 52.17 & 70.62 & 58.17 & 58.05 \\
DeMamba & 99.38 & 99.55 & \cellcolor{orange!24} 98.17 & \cellcolor{orange!24} 97.84 & \cellcolor{orange!24} 99.84 & 99.29 & \cellcolor{orange!24} 98.57 & 96.40 & 85.84 & 97.21 \\
Ours & \cellcolor{cyan!24} 99.96 & \cellcolor{cyan!24} 100.00 & \cellcolor{cyan!24} 100.00 & \cellcolor{cyan!24} 100.00 & \cellcolor{cyan!24} 100.00 & \cellcolor{cyan!24} 99.89 & \cellcolor{cyan!24} 100.00 & \cellcolor{cyan!24} 99.99 & \cellcolor{orange!24} 93.69 & \cellcolor{cyan!24} 99.28 \\
\midrule
\midrule
D3 & 84.87&89.21 &77.27 &70.47 &88.31 &77.19 &81.92 &52.08 &66.50 &76.42 \\
STALL & 47.77 & 58.16 & 36.83 & 23.19 & 33.70 & 35.68 & 31.06 & 11.17 & 26.29 & 33.76 \\
Over-Coherence & 6.34 & 5.39 & 6.44 & 7.48 & 5.80 & 6.25 & 5.77 & 1.70 & 5.86 & 5.67 \\
ReConFuse & 93.19 & 94.76 & 84.82 & 74.38 & 92.79 & 84.44 & 78.36 & 36.96 & 63.94 & 78.18 \\
DIVID-CNN & 85.74 & 80.09 & 59.18 & 55.47 & 93.80 & 79.48 & 38.08 & 43.94 & 42.99 & 64.31 \\
DIVID-LSTM & 73.85 & 52.20 & 40.20 & 29.18 & 80.41 & 58.55 & 27.05 & 14.22 & 29.67 & 45.04 \\
G2VD-CLIP & 94.81 & 92.90 & 76.70 & 81.54 & 97.96 & 94.87 & 60.33 & 61.40 & 56.90 & 79.71 \\
Ivy-xDetector & 82.57 & 77.70 & 40.16 & 48.25 & 88.90 & 64.45 & 31.43 & 37.77 & 67.23 & 59.83 \\
DeepTraceReward & \cellcolor{orange!24} 97.75 & \cellcolor{orange!24} 99.73 & 58.15 & \cellcolor{orange!24} 88.39 & 96.89 & 95.58 & 54.11 & \cellcolor{orange!24} 91.27 & \cellcolor{orange!24} 74.11 & 84.00 \\
MM-Det & 10.57 & 23.95 & 11.71 & 12.89 & 26.23 & 14.08 & 10.49 & 30.19 & 17.31 & 17.49 \\
DeMamba & 95.94 & 96.38 & \cellcolor{orange!24} 87.73 & 87.70 & \cellcolor{orange!24} 99.05 & \cellcolor{orange!24} 96.36 & \cellcolor{orange!24} 90.40 & 59.58 & 65.55 & \cellcolor{orange!24} 86.52 \\
Ours & \cellcolor{cyan!24} 99.73 & \cellcolor{cyan!24} 100.00 & \cellcolor{cyan!24} 100.00 & \cellcolor{cyan!24} 99.97 & \cellcolor{cyan!24} 100.00 & \cellcolor{cyan!24} 99.76 & \cellcolor{cyan!24} 100.00 & \cellcolor{cyan!24} 99.00 & \cellcolor{cyan!24} 88.23 & \cellcolor{cyan!24} 98.52 \\
\bottomrule
\end{tabular}
\end{adjustbox}
\label{tab:genvideo}
\end{table}

\begin{table}[t]
\caption{AUC and AP on the FakeParts dataset.}
\centering
\begin{adjustbox}{max width=\textwidth}
\begin{tabular}{lcccccccccc}
\toprule
\multirow{2}{*}{Model} & \multicolumn{2}{c}{CogVideoX} & \multicolumn{2}{c}{LTX-Video} & \multicolumn{2}{c}{AllegroAI} & \multicolumn{2}{c}{Veo2} & \multicolumn{2}{c}{Avg.} \\
\cmidrule(lr){2-3} \cmidrule(lr){4-5} \cmidrule(lr){6-7} \cmidrule(lr){8-9} \cmidrule(lr){10-11}
 & AUC & AP & AUC & AP & AUC & AP & AUC & AP & AUC & AP \\
\midrule
D3 & 93.60 & 93.86 & 91.13 & 90.58 & 83.08 & 79.26 & 79.41 & 72.27 & 86.81 & 83.99 \\
STALL & 83.68 & 83.80 & 67.77 & 69.09 & 68.99 & 70.67 & 60.01 & 58.54 & 70.11 & 70.53 \\
Over-Coherence & 41.14 & 43.83 & 45.45 & 43.04 & 37.13 & 40.02 & 35.94 & 39.63 & 39.92 & 41.63 \\
ReConFuse & 95.13 & 92.76 & 85.30 & 86.53 & 96.64 & 96.69 & 89.73 & 89.65 & 91.70 & 91.41 \\
DIVID-CNN & 58.75 & 65.11 & 89.72 & 89.60 & 98.21 & 98.20 & 90.15 & 90.38 & 84.21 & 85.82 \\
DIVID-LSTM & 59.63 & 57.25 & 82.66 & 86.71 & \cellcolor{orange!24} 98.82 & \cellcolor{orange!24} 98.93 & 88.96 & 88.77 & 82.52 & 82.92 \\
G2VD-CLIP & \cellcolor{orange!24} 97.13 & \cellcolor{orange!24} 96.57 & 91.01 & 90.72 & \cellcolor{cyan!24} 99.65 & \cellcolor{cyan!24} 99.64 & \cellcolor{orange!24} 96.61 & \cellcolor{orange!24} 95.29 & \cellcolor{orange!24} 96.10 & \cellcolor{orange!24} 95.56 \\
Ivy-xDetector & 96.62 & 95.95 & \cellcolor{orange!24} 94.52 & \cellcolor{cyan!24} 94.54 & 93.66 & 93.00 & 90.05 & 88.47 & 93.71 & 92.99 \\
DeepTraceReward & 74.11 & 70.49 & 87.03 & 84.20 & 92.98 & 93.21 & 91.23 & 91.59 & 86.34 & 84.87 \\
MM-Det & 53.18 & 51.70 & 55.53 & 55.10 & 60.26 & 56.10 & 54.99 & 53.04 & 55.99 & 53.99 \\
DeMamba & 96.70 & 94.57 & 86.96 & 87.62 & 98.62 & 98.47 & 92.75 & 91.19 & 93.76 & 92.96 \\
\midrule
Ours & \cellcolor{cyan!24} \revise{$99.17{\scriptscriptstyle\pm0.07}$} & \cellcolor{cyan!24} \revise{$99.07{\scriptscriptstyle\pm0.09}$} & \cellcolor{cyan!24} \revise{$95.56{\scriptscriptstyle\pm1.13}$} & \cellcolor{orange!24} \revise{$92.87{\scriptscriptstyle\pm0.58}$} & \revise{$98.67{\scriptscriptstyle\pm0.57}$} & \revise{$98.10{\scriptscriptstyle\pm0.66}$} & \cellcolor{cyan!24} \revise{$96.74{\scriptscriptstyle\pm0.69}$} & \cellcolor{cyan!24} \revise{$96.91{\scriptscriptstyle\pm1.47}$} & \cellcolor{cyan!24} \revise{$97.54{\scriptscriptstyle\pm0.38}$} & \cellcolor{cyan!24} \revise{$96.74{\scriptscriptstyle\pm0.53}$} \\
\bottomrule
\end{tabular}
\end{adjustbox}
\label{tab:fakeparts}
\end{table}

\subsection{Ablation study}
In this part, we conduct ablation studies to investigate the advantages of the proposed components: the adaptive reconstructor and reconstruction-error-based temporal modeling. 
We compare our method with \revise{four variants: (a) a method that uses a frozen VAE reconstructor with the same 3D classifier; (b) a method that uses a framewise 2D ResNet-18 with temporal feature averaging instead of the entire 3D classifier; (c) a method that explicitly uses only $\Delta_t^2 x_t-\Delta_t^2\widetilde{x}_t$ as input to the 2D classifier; and (d) a method that concatenates $\Delta_t^2 x_t-\Delta_t^2\widetilde{x}_t$ and the reconstruction errors along time before the 3D classifier. Except for variant (a), all use an adaptive VAE. We conduct the ablation study on FakeParts and HRFVideo. Due to page limit, Table~\ref{tab:ablation} reports the average results, and Tables~\ref{tab:ablation_fakeparts} and~\ref{tab:ablation_hrfvideo} report the per-generator results. Our method achieves the best average AUC and AP on both datasets. }

\begin{table}[t]
\centering
\color{black}
\captionsetup{labelfont={color=black},textfont={color=black}}
\caption{Component ablations on FakeParts and HRFVideo.} \label{tab:ablation}
\begin{adjustbox}{max width=\textwidth,center}\normalsize
\begin{tabular}{@{}lcccc@{}}
\toprule
\multirow{2}{*}{Variant} & \multicolumn{2}{c}{FakeParts} & \multicolumn{2}{c}{HRFVideo} \\
\cmidrule(lr){2-3}\cmidrule(l){4-5}
 & Avg. AUC & Avg. AP & Avg. AUC & Avg. AP \\
\midrule
(a) Frozen+Errors+Conv3D & $92.73{\scriptscriptstyle\pm0.12}$ & $92.04{\scriptscriptstyle\pm0.36}$ & $80.77{\scriptscriptstyle\pm0.28}$ & $61.84{\scriptscriptstyle\pm0.66}$ \\
(b) Adaptive+Errors+Conv2D & $95.93{\scriptscriptstyle\pm0.30}$ & $94.36{\scriptscriptstyle\pm0.27}$ & $90.96{\scriptscriptstyle\pm0.37}$ & $77.04{\scriptscriptstyle\pm0.35}$ \\
(c) Adaptive+$\Delta_t^2 x_t-\Delta_t^2\widetilde{x}_t$+Conv2D & $95.55{\scriptscriptstyle\pm0.39}$ & $93.26{\scriptscriptstyle\pm0.14}$ & $90.07{\scriptscriptstyle\pm0.23}$ & $76.86{\scriptscriptstyle\pm0.32}$ \\
(d) Adaptive+Errors\;\&\;$\Delta_t^2 x_t-\Delta_t^2\widetilde{x}_t$+Conv3D & \cellcolor{orange!24} $95.94{\scriptscriptstyle\pm0.12}$ & \cellcolor{orange!24} $95.02{\scriptscriptstyle\pm0.09}$ & \cellcolor{orange!24} $92.35{\scriptscriptstyle\pm0.39}$ & \cellcolor{orange!24} $82.21{\scriptscriptstyle\pm0.26}$ \\
\midrule
(Ours) Adaptive+Errors+Conv3D & \cellcolor{cyan!24} $97.54{\scriptscriptstyle\pm0.38}$ & \cellcolor{cyan!24} $96.74{\scriptscriptstyle\pm0.53}$ & \cellcolor{cyan!24} $94.36{\scriptscriptstyle\pm0.12}$ & \cellcolor{cyan!24} $82.46{\scriptscriptstyle\pm0.89}$ \\
\bottomrule
\end{tabular}\end{adjustbox}
\end{table}
\revise{Compared with the frozen-VAE variant (a), our method improves
average AUC by 4.81 and 13.59 percentage points on FakeParts
and HRFVideo, respectively. 
These gains show that learned reconstruction errors better support real/fake discrimination.
Compared with the framewise 2D variant (b), our method improves average AUC by 1.61 and 3.40 points. Hence, the 3D convolution extracts useful temporal evidence from the reconstruction errors themselves, without an additional raw-video branch. Next, variant (c) is 1.99 and 4.29 points below ours. Note that, variant (c) fixes the temporal coefficients to $(1,-2,1)$, whereas our temporal coefficients are learned; the gap shows that the learned temporal features are more discriminative. Finally, variant (d) is 1.60 and 2.01 points below ours. The extra $\Delta_t^2 x_t-\Delta_t^2\widetilde{x}_t$ slice does not enlarge the function class and only increases the input dimension, so it brings no benefit. This supports applying the 3D convolution directly to the reconstruction errors.}

\subsection{Analysis of Training Configurations}

We perform an analysis of the proposed detector with respect to two important training configurations: (i) the number of epochs and (ii) the LoRA rank. The analysis is conducted on the FakeParts dataset. When studying the number of epochs, the LoRA rank is fixed; the same applies when studying the LoRA rank. For each configuration, we consider 3 possible values.

\begin{figure}[!tb]
\centering
\includegraphics[width=0.85\linewidth]{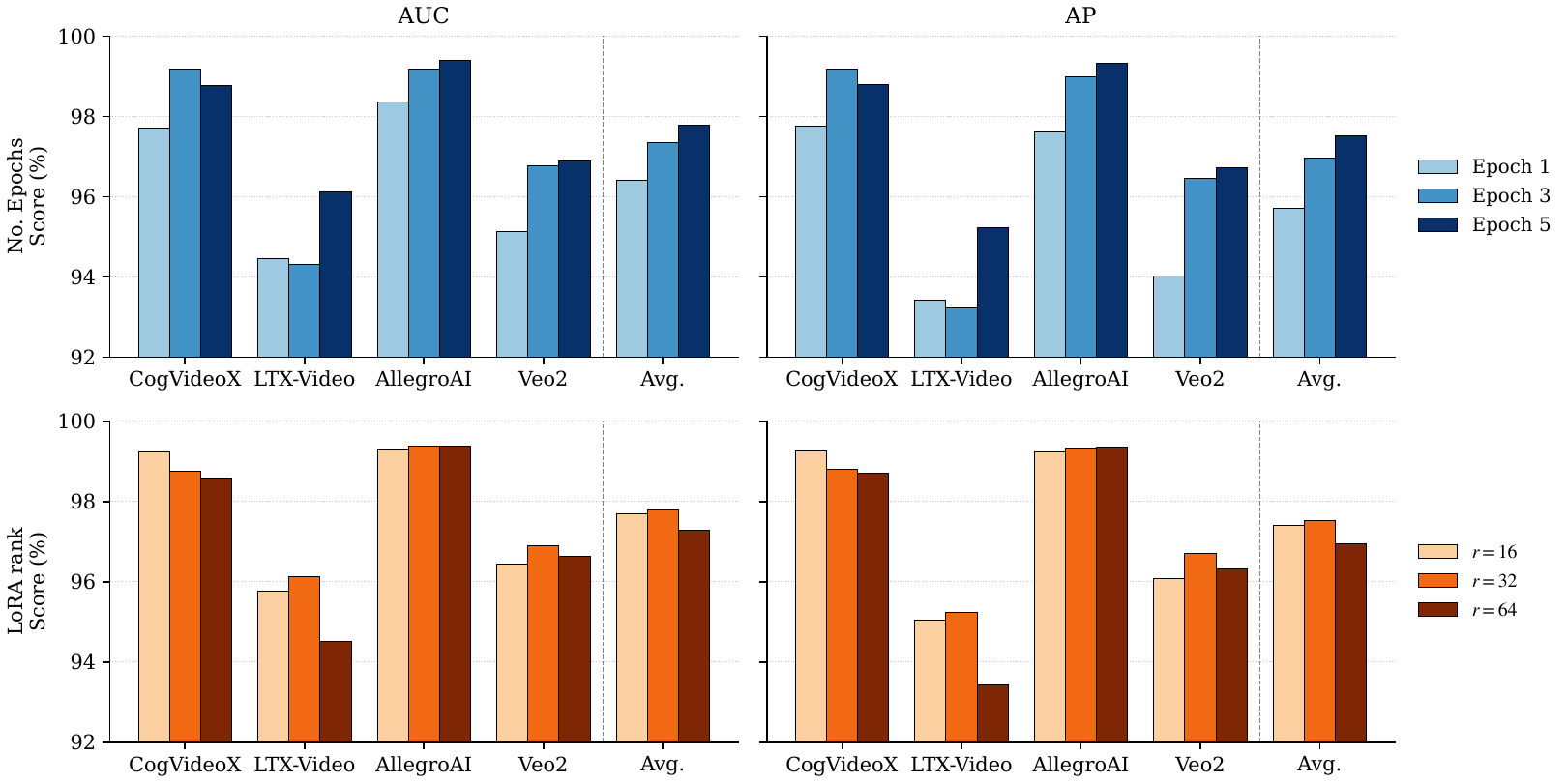}
\vspace*{-8pt}
\caption{Sensitivity to the number of epochs (upper panels) and LoRA rank (lower panels). Left panels are AUC, while the right one are AP. The default training epochs is selected on the validation set and the LoRA rank is $r=32$.}\label{fig:sensitivity}
\vspace*{-12pt}
\end{figure}

Figure~\ref{fig:sensitivity} shows that performance is relatively stable across the number of epochs, with more training epochs generally providing better overall performance, although the gains diminish at later epochs. Empirically, the final number of epochs is selected based on detection performance on a standalone validation set, and we find that this choice already yields desirable performance. As for the LoRA rank, we compare three LoRA ranks $r\in\{16,32,64\}$. We can see that, overall, the value of $r$ has only a marginal impact on detection performance---the variation in average AUC/AP is less than $\pm 1.0$. This suggests that the method is not sensitive to $r$. 

\subsection{\revise{Robustness experiments}}
\revise{\textbf{Robustness to video compression.} In practice, the video to be detected may be compressed. This part studies whether the proposal is robust to video compression. Specifically, we apply identical compression to real and fake videos, and evaluate the pretrained checkpoint. Two kinds of compression are considered: (i) \emph{video-level compression}: source videos are re-encoded with H.264 at constant rate factors (CRFs) of 18, 23, and 28 before frame extraction, where a larger CRF indicates stronger compression; (ii) \emph{frame-level processing}: the extracted frames undergo (a) JPEG compression, (b) downsampling of their width and height followed by restoration to the original dimensions, or (c) a combination of both. Results on FakeParts and HRFVideo are reported in Tables~\ref{tab:postprocessing_fakeparts} and~\ref{tab:postprocessing_hrfvideo}, respectively.
\begin{table}[!tb]
\centering
\caption{\revise{Robustness under video compression settings on FakeParts. Each generator cell reports AUC/AP (\%).}}
\label{tab:postprocessing_fakeparts}
\revise{\begin{minipage}{\textwidth}
\centering\footnotesize
\setlength{\tabcolsep}{2pt}
\renewcommand{\arraystretch}{1.12}
\begin{tabular*}{\textwidth}{@{\extracolsep{\fill}}llccccc@{}}
\toprule
Level & Compression & CogVideoX & LTX-Video & AllegroAI & Veo2 & Avg. \\
\midrule
\multirow{3}{*}{Video} & H.264 CRF 18 & 98.92/98.91 & 93.31/92.09 & 99.18/98.85 & 96.32/95.74 & 96.93/96.40 \\
& H.264 CRF 23 & 99.18/99.21 & 93.75/92.74 & 99.28/99.00 & 96.04/95.62 & 97.06/96.64 \\
& H.264 CRF 28 & 99.27/99.25 & 94.35/93.28 & 99.28/99.11 & 94.91/94.32 & 96.95/96.49 \\
\midrule
\multirow{3}{*}{Frame} & JPEG & 99.17/98.93 & 96.03/95.01 & 99.34/99.01 & 97.46/97.30 & 98.00/97.56 \\
& Downsample & 98.79/98.82 & 92.32/91.47 & 99.01/98.68 & 96.01/95.50 & 96.53/96.12 \\
& JPEG + Downsample & 98.82/98.58 & 94.23/93.24 & 99.14/98.68 & 96.66/96.35 & 97.21/96.71 \\
 \midrule
\multicolumn{2}{@{}c}{No Compression} & 99.11/99.11 & 94.32/93.17 & 99.18/98.87 & 96.77/96.47 & 97.35/96.91 \\
\bottomrule
\end{tabular*}
\end{minipage}}
\end{table}

Under video-level compression, the performance on FakeParts is nearly unchanged, while that on HRFVideo decreases gradually as the CRF increases, with the average AUC remaining above 92\% even at CRF 28. Frame-level processing has only a marginal effect on both datasets, and some settings even slightly improve the performance. These results suggest that the detection performance is not primarily driven by compression, and that the learned reconstruction errors retain discriminative evidence after common video compression.}

\revise{\textbf{Robustness to temporal sampling.}
The temporal sampling rate determines the time interval between the three consecutive reconstruction errors covered by a temporal kernel. As in many existing methods (e.g., D3), the sampling rate is a hyperparameter of our method. Our default implementation samples frames at 2 fps (Appendix~\ref{sec:implement-details}), which was not tuned for detection performance; instead, we chose a low rate so that nine frames cover about four seconds, which keeps the number of processed frames and thus the computational cost small. To examine the sensitivity to this choice, we retrain and evaluate the detector at 4 and 8 fps under otherwise identical settings and compare it with the default 2 fps on the FakeParts dataset. Table~\ref{tab:temporal_sampling} reports the per-generator results. The sampling rate has only a marginal impact on AUC and AP, with differences within $\pm 1$ point, suggesting that the detector is not sensitive to the choice of sampling rate.}

{\color{black}
\vspace*{-6pt}
\subsection{Computational analysis}
\vspace*{-6pt}
This part studies the inference cost of the proposal. Specifically, we measure the floating-point operations (FLOPs) and the latency for processing one video on a A100-80GB GPU; for our method, FLOPs includes the VAE encoding and decoding as well as the classifier, so does latency. Table~\ref{tab:inference_cost} shows our method takes only 0.117\,s per video on average, which is acceptable for processing large-scale benchmark datasets. This latency is competitive with the training-based baselines and is lower than that of most of them. The training cost is discussed in Appendix~\ref{sec:implement-details}.}

\vspace*{-6pt}
{\color{black}\section{Conclusion and Discussion}
\vspace*{-6pt}
In this paper, we develop an algorithm that produces discriminative reconstruction errors and exploits the temporal signal to detect AI-generated videos. The overall workflow is simple and easy to implement. Across the evaluated benchmarks, the method achieves competitive performance against the considered training-based and training-free detectors, and the ablation study confirms the effectiveness of its key components.

Like existing detectors, our method makes only video-level decisions, whereas some forgeries, such as the inpainted videos in FakeParts, alter only part of a real video. A natural next step is to localize the AI-manipulated regions and frames in human-AI co-generated videos.}

\section*{AI Use Statement}
In preparing this paper, LLMs were used only to aid and polish the writing and to retrieve and discover related work. All cited references were manually checked by the authors to confirm that the cited works exist. This use did not affect the core methodology.

\section*{Ethics Statement}
This work uses only synthetic and existing public benchmark datasets and involves no participant recruitment or new collection of personal data. 

\section*{Reproducibility Statement}
Section~\ref{sec:method} in the main text describes the proposed method and the training and evaluation protocols. \revise{Section~\ref{sec:experiments} provides a high-level experimental design. Appendix~\ref{sec:implement-details}--\ref{sec:experiments-details} document the model architecture, preprocessing, optimization settings, baseline implementation details, data sources, and hardware and software configuration.}

\bibliographystyle{iclr2027_conference}
\bibliography{iclr2027_conference}

\newpage
\appendix

\setcounter{table}{0}
\setcounter{figure}{0}
\renewcommand{\thefigure}{A\arabic{figure}}
\renewcommand{\thetable}{A\arabic{table}}

\section{Additional related works}\label{sec:addition-related-work}
\textbf{AI Video Generation}. Modern AI video generation is largely built upon three broad methodological paradigms: diffusion-based, flow-matching, and autoregressive models. Diffusion-based generative frameworks~\citep{ho2022videodiffusion, blattmann2023align} are widely used for video generation. Building on these frameworks, subsequent studies have developed models for high-quality text-to-video and image-to-video generation~\citep{he2022latent, ho2022imagen, singer2022make, blattmann2023svd, guo2023animatediff, chen2024videocrafter2, wang2025lavie}. Furthermore, diffusion transformers provide a scalable backbone for image generation~\citep{peebles2023scalable}. Related Transformer-based architectures have also been developed for text-to-image generation~\citep{esser2024scaling}. Subsequent studies have extended and refined Transformer-based video generation~\citep{gupta2024photorealistic, yang2024cogvideox, zheng2024open, ma2025latte, zhao2025real, chen2026sanavideo, jia2026moga}. More recently, flow matching~\citep{lipman2023flowmatching} has emerged as a new approach for video generation~\citep{jin2025pyramidal, ifriqi2026flowception, kodaira2026streamdit, pan2026temporal} and has shown competitive performance. Autoregressive methods~\citep{yan2021videogpt, villegas2022phenaki, dan2024videopoet} generate video tokens or segments conditioned on previously generated content. Recent autoregressive diffusion models further support long-form and streaming video generation~\citep{yang2025longlive, yin2025slow, huang2026self, liu2026rolling}. Such diversity in generative formulations and architectures may produce heterogeneous spatiotemporal artifacts, posing challenges for generalization to unseen video generators.

\section{Implementation Details}\label{sec:implement-details}
\paragraph{Video preprocessing.}
We sample video frames at 2 frames per second, using $9$ frames when available and $5$ otherwise, with last-frame repetition for shorter clips. Random temporal windows are used during training and the central window during evaluation. Frames are resized while preserving their aspect ratio, with \revise{the total number of pixels per frame constrained to $[224^2,448^2]$} and both height and width adjusted to multiples of eight. No spatial cropping or padding is applied. \revise{Resizing uses bilinear interpolation with antialiasing. Videos with the same frame count, height, and width are grouped into the same batch.}

\paragraph{\revise{Reconstructor.}}
We use the pretrained Wan2.1 VAE\footnote{\url{https://huggingface.co/Wan-AI/Wan2.1-T2V-1.3B-Diffusers}} as the reconstruction backbone. \revise{Videos are reconstructed deterministically using the posterior mode. Inputs are scaled from $[0,1]$ to $[-1,1]$ to adapt the input form of Wan2.1 VAE. Following the setup of Wan2.1 VAE, we retain the output clipping, and we map reconstructions back to $[0,1]$ before computing reconstruction errors. We do not perform per-video normalization.} The pretrained Wan-VAE parameters are frozen, and rank-$32$ LoRA adapters are inserted into eight 3D convolutional layers in the encoder and decoder middle blocks of the VAE.

\paragraph{\revise{Classifier.}}
The reconstruction error is processed by a randomly initialized 3D ResNet-18-style classifier with stage depths $(2,2,2,2)$ and channel widths $(32,64,128,256)$. \revise{Following the kernel configuration of \texttt{torchvision.models.video.resnet}\footnote{\url{https://docs.pytorch.org/vision/stable/_modules/torchvision/models/video/resnet.html}}, the stem uses a $3\times7\times7$ kernel, while residual blocks use $3\times3\times3$ kernels. The classifier uses symmetric zero-padding, group normalization with 32 groups, and ReLU as the activation function.} All downsampling is spatial, with no temporal downsampling throughout the network. Global average pooling \revise{over time and space} of the final-stage feature map produces a 256-dimensional descriptor, followed by a $256\!\rightarrow\!512\!\rightarrow\!256\!\rightarrow\!1$ classification head. \revise{Each hidden linear layer is preceded by LayerNorm and followed by GELU and dropout with probability $0.1$.}

\paragraph{\revise{Optimization.}}
The LoRA adapters and detector are jointly optimized for five epochs using AdamW \citep{loshchilov2018decoupled} with weight decay $10^{-4}$. The initial learning rates are $5\times10^{-6}$ for LoRA and $5\times10^{-5}$ for the detector, with cosine decay to $10^{-7}$. Training uses a global batch size of eight. The LoRA adapters are optimized solely through the video-level real/fake classification objective. \revise{We did not use any kind of data augmentation.} Training and evaluation use BF16 automatic mixed precision. 

\paragraph{\revise{Training cost.}}
\revise{For our method, only the LoRA adapters and the classifier, 11.32M of the 138.21M parameters, are trained. A three-epoch run on the full training videos takes 24.97\,h on two A100 GPUs, i.e., the training can be completed in about one day. Once trained, the detector is directly applied to new videos without retraining.}

\section{Experimental Details}\label{sec:experiments-details}

\paragraph{Implementation baselines.} For the training-based methods, we directly reuse their checkpoints whenever they are available. These methods include  Ivy-xDetector\footnote{\url{https://huggingface.co/AI-Safeguard/Ivy-Fake}}, DeepTraceReward\footnote{\url{https://huggingface.co/DeeptraceReward/RewardModel}}, and MM-Det\footnote{\url{https://huggingface.co/sparklexfantasy/llava-7b-1.5-rfrd}}. When checkpoints are unavailable, we implement these methods ourselves or reuse their training code. These methods include ReConFuse, two variants of DIVID, G2VD, and DeMamba\footnote{\url{https://github.com/chenhaoxing/DeMamba}}. For the training-free methods, we use their official implementations, including D3\footnote{\url{https://github.com/Zig-HS/D3}}, STALL\footnote{\url{https://github.com/OmerBenHayun/STALL}}, and Over-Coherence\footnote{\url{https://github.com/FujitsuResearch/training-free-detection-of-text-to-video-generations-via-over-coherence}}.

\paragraph{Data sources.}
Kinetics-400 videos\footnote{\url{https://github.com/cvdfoundation/kinetics-dataset}}
and released VBench generations\footnote{\url{https://github.com/Vchitect/VBench}}
are obtained from their public repositories.
MovieGen and Panda-70M videos are obtained from the
MovieGenBench\footnote{\url{https://github.com/facebookresearch/MovieGenBench}}
and Panda-70M\footnote{\url{https://github.com/snap-research/Panda-70M}}
releases, respectively.
For GenVideo, we use the released \texttt{GenVideo-Val} split,
following the download and preparation instructions in the
DeMamba repository.\footnote{\url{https://github.com/chenhaoxing/DeMamba}}
EvalCrafter\footnote{\url{https://github.com/EvalCrafter/EvalCrafter}}
and VideoPhy\footnote{\url{https://github.com/Hritikbansal/videophy}}
videos are obtained from their respective public releases.
FakeParts videos and metadata are obtained from its
Hugging Face release.\footnote{\url{https://huggingface.co/datasets/hi-paris/FakeParts}}

\paragraph{Hardware setup.}
\revise{All experiments are conducted on (i) NVIDIA GH200 Grace Hopper GPUs, each with 96\,GB of GPU memory and (ii) NVIDIA A100, with GPU memory 80\,GB.} Our method is trained on two GPUs using distributed data parallelism. The version of Python is 3.12 with \texttt{PyTorch 2.8.0a0}, and CUDA~12.9.

\section{Additional Results}

\begin{table}[H]
\centering
\caption{AUC and AP on classifying the fake videos in the VideoPhy dataset and real videos in the MSR-VTT dataset. AUC is demonstrated on the upper part while AP is on the lower part.}
\begin{adjustbox}{width=\textwidth}
\begin{tabular}{lcccccccccc}
\toprule
Model & {LaVIE} & {cogvideox} & {cogvideox-5B} & {dream\_machine} & {Gen-2} & {pika\_video} & {svd} & {vc2} & {zeroscope} & Avg. \\
\midrule
D3 & 52.11 & 60.96 & 60.40 & 70.80 & 88.20 & 81.99 & 74.11 & 69.03 & 69.31 & 69.66 \\
STALL & 69.85 & 60.39 & 59.21 & 68.55 & 84.29 & 76.78 & 71.24 & 81.63 & 69.16 & 71.23 \\
Over-Coherence & 21.70 & 39.08 & 35.63 & 62.70 & 52.51 & 29.19 & 35.59 & 44.59 & 38.12 & 39.90 \\
ReConFuse & 96.30 & 92.85 & 94.27 & 98.23 & 99.65 & 99.32 & 97.94 & 99.41 & 95.52 & 97.05 \\
DIVID-CNN & 96.45 & 79.53 & 88.62 & 97.43 & 99.00 & 99.39 & 97.30 & 99.47 & 81.14 & 93.15 \\
DIVID-LSTM & 92.46 & 73.77 & 84.42 & 95.33 & 98.21 & 99.10 & 94.58 & 98.69 & 67.67 & 89.36 \\
G2VD-CLIP & \cellcolor{orange!24} 99.04 & 90.37 & 92.68 & \cellcolor{orange!24} 99.47 & \cellcolor{orange!24} 99.83 & 99.70 & 99.20 & 99.81 & 96.84 & 97.44 \\
Ivy-xDetector & 83.95 & 86.36 & 83.71 & 94.26 & 98.73 & 96.03 & 94.80 & 89.28 & 84.01 & 90.13 \\
DeepTraceReward & 70.98 & 88.08 & 77.03 & 95.45 & 99.32 & 98.59 & 94.68 & 78.32 & 77.67 & 86.68 \\
MM-Det & 50.03 & 52.60 & 57.87 & 52.34 & 55.69 & 50.59 & 56.52 & 51.56 & 51.42 & 53.18 \\
DeMamba & 98.68 & \cellcolor{orange!24} 97.00 & \cellcolor{orange!24} 97.96 & 99.38 & 99.76 & \cellcolor{orange!24} 99.89 & \cellcolor{orange!24} 99.49 & \cellcolor{orange!24} 99.91 & \cellcolor{orange!24} 98.01 & \cellcolor{orange!24} 98.90 \\
Ours & \cellcolor{cyan!24} 99.86 & \cellcolor{cyan!24} 100.00 & \cellcolor{cyan!24} 99.99 & \cellcolor{cyan!24} 100.00 & \cellcolor{cyan!24} 100.00 & \cellcolor{cyan!24} 100.00 & \cellcolor{cyan!24} 99.55 & \cellcolor{cyan!24} 100.00 & \cellcolor{cyan!24} 99.90 & \cellcolor{cyan!24} 99.92 \\
\midrule
\midrule
D3 & 15.78 & 26.12 & 18.43 & 22.66 & 41.03 & 33.92 & 25.57 & 23.07 & 22.01 & 25.40 \\
STALL & 25.81 & 18.91 & 19.87 & 22.86 & 46.09 & 33.56 & 29.09 & 40.88 & 23.00 & 28.90 \\
Over-Coherence & 8.92 & 13.72 & 10.75 & 17.92 & 13.34 & 8.68 & 10.66 & 12.66 & 15.25 & 12.43 \\
ReConFuse & 84.40 & 69.38 & 77.73 & 91.93 & 97.80 & 95.76 & 91.12 & 97.49 & 78.49 & 87.12 \\
DIVID-CNN & 86.55 & 54.67 & 70.72 & 91.70 & 96.62 & 97.27 & 91.96 & 97.97 & 50.44 & 81.99 \\
DIVID-LSTM & 75.52 & 38.34 & 53.24 & 82.36 & 93.29 & 95.37 & 83.00 & 94.94 & 30.15 & 71.80 \\
G2VD-CLIP & \cellcolor{orange!24} 95.32 & 71.10 & 76.06 & 96.27 & \cellcolor{orange!24} 98.57 & 97.34 & 96.42 & 98.90 & 86.39 & 90.71 \\
Ivy-xDetector & 46.20 & 47.50 & 41.35 & 70.66 & 91.26 & 72.63 & 73.48 & 53.10 & 37.54 & 59.30 \\
DeepTraceReward & 35.60 & 68.42 & 47.38 & 91.78 & 98.23 & 94.69 & 89.16 & 39.78 & 40.51 & 67.28 \\
MM-Det & 14.65 & 16.29 & 23.16 & 15.98 & 18.46 & 13.33 & 21.02 & 15.27 & 15.29 & 17.05 \\
DeMamba & 91.66 & \cellcolor{orange!24} 81.41 & \cellcolor{orange!24} 89.06 & \cellcolor{orange!24} 96.56 & 98.36 & \cellcolor{orange!24} 99.45 & \cellcolor{orange!24} 97.76 & \cellcolor{orange!24} 99.46 & \cellcolor{orange!24} 87.08 & \cellcolor{orange!24} 93.42 \\
\midrule
Ours & \cellcolor{cyan!24} 99.70 & \cellcolor{cyan!24} 100.00 & \cellcolor{cyan!24} 99.97 & \cellcolor{cyan!24} 100.00 & \cellcolor{cyan!24} 100.00 & \cellcolor{cyan!24} 100.00 & \cellcolor{cyan!24} 98.93 & \cellcolor{cyan!24} 100.00 & \cellcolor{cyan!24} 99.45 & \cellcolor{cyan!24} 99.78 \\
\bottomrule
\end{tabular}
\end{adjustbox}
\label{tab:videophy-ap}
\end{table}

\begin{table}[H]
\centering
\scriptsize
\caption{\revise{AUC (upper part) and AP (lower part) across generators on HRFVideo dataset. }}
\label{tab:hrfvideo_reconstruction}
\begingroup\color{black}
\renewcommand{\arraystretch}{1.05}
\resizebox{\textwidth}{!}{%
\begin{tabular}{@{}lrrrrrr@{}}
\toprule
Model & Hailuo & Seaweed & Seedance & StepVideo & Wan2.1 & Avg. \\
\midrule
D3 & 59.74 & 67.61 & 62.86 & 72.22 & 61.10 & 64.71 \\
STALL & 73.57 & 80.25 & 80.27 & 75.17 & 79.57 & 77.77 \\
Over-Coherence & 54.89 & 79.90 & 88.17 & 76.57 & 76.37 & 75.18 \\
ReConFuse & 76.40 & 86.40 & 83.47 & 84.44 & 85.50 & 83.24 \\
DIVID-CNN & 75.70 & 91.22 & 80.07 & 80.62 & 78.90 & 81.30 \\
DIVID-LSTM & 77.43 & 93.90 & 72.24 & 85.47 & 82.81 & 82.37 \\
G2VD-CLIP & \cellcolor{orange!24} 83.14 & 87.33 & 81.41 & 72.41 & 88.16 & 82.49 \\
Ivy-xDetector & 65.58 & 72.86 & 70.53 & 54.60 & 75.12 & 67.74 \\
DeepTraceReward & 73.23 & 91.75 & 78.49 & 38.28 & 37.62 & 63.87 \\
MM-Det & 68.82 & 53.20 & 64.84 & 54.84 & 73.78 & 63.10 \\
DeMamba & 82.67 & \cellcolor{cyan!24} 95.88 & \cellcolor{orange!24} 90.10 & \cellcolor{orange!24} 87.49 & \cellcolor{orange!24} 97.43 & \cellcolor{orange!24} 90.71 \\
Ours & \cellcolor{cyan!24} \revise{$93.81{\scriptscriptstyle\pm0.23}$} & \cellcolor{orange!24} \revise{$94.16{\scriptscriptstyle\pm0.55}$} & \cellcolor{cyan!24} \revise{$95.78{\scriptscriptstyle\pm0.43}$} & \cellcolor{cyan!24} \revise{$90.12{\scriptscriptstyle\pm0.68}$} & \cellcolor{cyan!24} \revise{$97.94{\scriptscriptstyle\pm0.55}$} & \cellcolor{cyan!24} \revise{$94.36{\scriptscriptstyle\pm0.12}$} \\
\midrule
\midrule
D3 & 36.28 & 42.23 & 39.38 & 46.31 & 42.05 & 41.25 \\
STALL & 46.92 & 55.08 & 60.78 & 45.72 & 55.42 & 52.78 \\
Over-Coherence & 30.67 & 58.49 & 75.50 & 52.23 & 59.39 & 55.26 \\
ReConFuse & 58.72 & 75.39 & 68.38 & \cellcolor{orange!24} 74.25 & 74.99 & 70.35 \\
DIVID-CNN & 53.93 & 78.55 & 60.93 & 63.52 & 61.11 & 63.61 \\
DIVID-LSTM & 53.61 & \cellcolor{orange!24} 87.38 & 49.58 & 72.35 & 64.32 & 65.45 \\
G2VD-CLIP & \cellcolor{orange!24} 67.05 & 71.45 & 63.81 & 57.96 & 74.36 & 66.93 \\
Ivy-xDetector & 43.30 & 51.41 & 49.63 & 30.76 & 53.72 & 45.76 \\
DeepTraceReward & 47.33 & 75.50 & 52.72 & 22.94 & 22.76 & 44.25 \\
MM-Det & 43.62 & 29.83 & 41.47 & 30.11 & 52.32 & 39.47 \\
DeMamba & 65.99 & \cellcolor{cyan!24} 91.74 & \cellcolor{orange!24} 81.25 & \cellcolor{cyan!24} 76.04 & \cellcolor{cyan!24} 94.05 & \cellcolor{orange!24} 81.81 \\
Ours & \cellcolor{cyan!24} \revise{$82.89{\scriptscriptstyle\pm0.22}$} & \revise{$82.72{\scriptscriptstyle\pm1.99}$} & \cellcolor{cyan!24} \revise{$87.95{\scriptscriptstyle\pm1.10}$} & \revise{$65.03{\scriptscriptstyle\pm0.81}$} & \cellcolor{orange!24} \revise{$93.72{\scriptscriptstyle\pm2.27}$} & \cellcolor{cyan!24} \revise{$82.46{\scriptscriptstyle\pm0.89}$} \\
\bottomrule
\end{tabular}%
}
\endgroup
\end{table}

\begin{table}[H]
\centering
\color{black}
\captionsetup{labelfont={color=black},textfont={color=black}}
\caption{Ablation studies on FakeParts by generator. AUC is reported in the upper part and AP in the lower part.}
\label{tab:ablation_fakeparts}
\begin{adjustbox}{max width=\textwidth}
\begin{tabular}{lccccc}
\toprule
Variant & CogVideoX & LTX-Video & AllegroAI & Veo2 & Avg. \\
\midrule
(a) Frozen+Errors+Conv3D & $93.28{\scriptscriptstyle\pm0.47}$ & \cellcolor{orange!24} $94.20{\scriptscriptstyle\pm0.18}$ & $94.17{\scriptscriptstyle\pm0.66}$ & $89.25{\scriptscriptstyle\pm0.56}$ & $92.73{\scriptscriptstyle\pm0.12}$ \\
(b) Adaptive+Errors+Conv2D & $98.55{\scriptscriptstyle\pm0.56}$ & $92.36{\scriptscriptstyle\pm0.08}$ & $98.14{\scriptscriptstyle\pm0.25}$ & \cellcolor{orange!24} $94.66{\scriptscriptstyle\pm0.83}$ & $95.93{\scriptscriptstyle\pm0.30}$ \\
(c) Adaptive+$\Delta_t^2 x_t-\Delta_t^2\widetilde{x}_t$+Conv2D & \cellcolor{cyan!24} $99.57{\scriptscriptstyle\pm0.32}$ & $91.98{\scriptscriptstyle\pm0.56}$ & $97.30{\scriptscriptstyle\pm0.65}$ & $93.33{\scriptscriptstyle\pm0.79}$ & $95.55{\scriptscriptstyle\pm0.39}$ \\
(d) Adaptive+Errors\&$\Delta_t^2 x_t-\Delta_t^2\widetilde{x}_t$+Conv3D & \cellcolor{orange!24} $99.35{\scriptscriptstyle\pm0.41}$ & $91.95{\scriptscriptstyle\pm0.20}$ & \cellcolor{orange!24} $98.24{\scriptscriptstyle\pm0.15}$ & $94.20{\scriptscriptstyle\pm0.46}$ & \cellcolor{orange!24} $95.94{\scriptscriptstyle\pm0.12}$ \\
\midrule
(Ours) Adaptive+Errors+Conv3D & $99.17{\scriptscriptstyle\pm0.07}$ & \cellcolor{cyan!24} $95.56{\scriptscriptstyle\pm1.13}$ & \cellcolor{cyan!24} $98.67{\scriptscriptstyle\pm0.57}$ & \cellcolor{cyan!24} $96.74{\scriptscriptstyle\pm0.69}$ & \cellcolor{cyan!24} $97.54{\scriptscriptstyle\pm0.38}$ \\
\midrule
\midrule
(a) Frozen+Errors+Conv3D & $92.21{\scriptscriptstyle\pm0.57}$ & \cellcolor{cyan!24} $93.34{\scriptscriptstyle\pm0.15}$ & $93.99{\scriptscriptstyle\pm0.48}$ & $88.61{\scriptscriptstyle\pm0.75}$ & $92.04{\scriptscriptstyle\pm0.36}$ \\
(b) Adaptive+Errors+Conv2D & $98.61{\scriptscriptstyle\pm0.33}$ & $88.71{\scriptscriptstyle\pm0.45}$ & $97.28{\scriptscriptstyle\pm0.42}$ & $92.85{\scriptscriptstyle\pm0.26}$ & $94.36{\scriptscriptstyle\pm0.27}$ \\
(c) Adaptive+$\Delta_t^2 x_t-\Delta_t^2\widetilde{x}_t$+Conv2D & \cellcolor{cyan!24} $99.46{\scriptscriptstyle\pm0.65}$ & $87.03{\scriptscriptstyle\pm0.76}$ & $95.61{\scriptscriptstyle\pm0.42}$ & $90.92{\scriptscriptstyle\pm0.20}$ & $93.26{\scriptscriptstyle\pm0.14}$ \\
(d) Adaptive+Errors\&$\Delta_t^2 x_t-\Delta_t^2\widetilde{x}_t$+Conv3D & \cellcolor{orange!24} $99.35{\scriptscriptstyle\pm0.47}$ & $89.56{\scriptscriptstyle\pm0.10}$ & \cellcolor{orange!24} $97.91{\scriptscriptstyle\pm0.67}$ & \cellcolor{orange!24} $93.27{\scriptscriptstyle\pm0.40}$ & \cellcolor{orange!24} $95.02{\scriptscriptstyle\pm0.09}$ \\
\midrule
(Ours) Adaptive+Errors+Conv3D & $99.07{\scriptscriptstyle\pm0.09}$ & \cellcolor{orange!24} $92.87{\scriptscriptstyle\pm0.58}$ & \cellcolor{cyan!24} $98.10{\scriptscriptstyle\pm0.66}$ & \cellcolor{cyan!24} $96.91{\scriptscriptstyle\pm1.47}$ & \cellcolor{cyan!24} $96.74{\scriptscriptstyle\pm0.53}$ \\
\bottomrule
\end{tabular}
\end{adjustbox}
\end{table}

\begin{table}[H]
\centering
\color{black}
\captionsetup{labelfont={color=black},textfont={color=black}}
\caption{Ablation studies on HRFVideo by generator. }
\label{tab:ablation_hrfvideo}
\begin{adjustbox}{max width=\textwidth}
\begin{tabular}{lcccccc}
\toprule
Variant & Wan2.1 & Hailuo & Seaweed & Seedance & StepVideo & Avg. \\
\midrule
(a) Frozen+Errors+Conv3D & $89.23{\scriptscriptstyle\pm0.75}$ & $76.00{\scriptscriptstyle\pm0.14}$ & $83.97{\scriptscriptstyle\pm0.74}$ & $83.74{\scriptscriptstyle\pm0.17}$ & $70.89{\scriptscriptstyle\pm0.39}$ & $80.77{\scriptscriptstyle\pm0.28}$ \\
(b) Adaptive+Errors+Conv2D & \cellcolor{orange!24} $96.99{\scriptscriptstyle\pm0.45}$ & $92.74{\scriptscriptstyle\pm0.80}$ & $90.51{\scriptscriptstyle\pm0.74}$ & $93.58{\scriptscriptstyle\pm0.93}$ & $80.97{\scriptscriptstyle\pm0.76}$ & $90.96{\scriptscriptstyle\pm0.37}$ \\
(c) Adaptive+$\Delta_t^2 x_t-\Delta_t^2\widetilde{x}_t$+Conv2D & $96.91{\scriptscriptstyle\pm0.19}$ & $93.36{\scriptscriptstyle\pm0.35}$ & $92.31{\scriptscriptstyle\pm0.48}$ & $94.56{\scriptscriptstyle\pm0.17}$ & $73.23{\scriptscriptstyle\pm0.26}$ & $90.07{\scriptscriptstyle\pm0.23}$ \\
(d) Adaptive+Errors\&$\Delta_t^2 x_t-\Delta_t^2\widetilde{x}_t$+Conv3D & $95.24{\scriptscriptstyle\pm0.69}$ & \cellcolor{cyan!24} $94.28{\scriptscriptstyle\pm0.40}$ & \cellcolor{cyan!24} $94.71{\scriptscriptstyle\pm0.67}$ & \cellcolor{orange!24} $95.71{\scriptscriptstyle\pm0.23}$ & \cellcolor{orange!24} $81.83{\scriptscriptstyle\pm0.77}$ & \cellcolor{orange!24} $92.35{\scriptscriptstyle\pm0.39}$ \\
\midrule
(Ours) Adaptive+Errors+Conv3D & \cellcolor{cyan!24} $97.94{\scriptscriptstyle\pm0.55}$ & \cellcolor{orange!24} $93.81{\scriptscriptstyle\pm0.23}$ & \cellcolor{orange!24} $94.16{\scriptscriptstyle\pm0.55}$ & \cellcolor{cyan!24} $95.78{\scriptscriptstyle\pm0.43}$ & \cellcolor{cyan!24} $90.12{\scriptscriptstyle\pm0.68}$ & \cellcolor{cyan!24} $94.36{\scriptscriptstyle\pm0.12}$ \\
\midrule
\midrule
(a) Frozen+Errors+Conv3D & $78.48{\scriptscriptstyle\pm0.45}$ & $54.91{\scriptscriptstyle\pm0.94}$ & $63.98{\scriptscriptstyle\pm0.77}$ & $69.25{\scriptscriptstyle\pm0.46}$ & $42.60{\scriptscriptstyle\pm0.89}$ & $61.84{\scriptscriptstyle\pm0.66}$ \\
(b) Adaptive+Errors+Conv2D & $90.98{\scriptscriptstyle\pm0.26}$ & $79.76{\scriptscriptstyle\pm0.62}$ & $73.27{\scriptscriptstyle\pm0.37}$ & $81.05{\scriptscriptstyle\pm0.86}$ & $60.15{\scriptscriptstyle\pm0.23}$ & $77.04{\scriptscriptstyle\pm0.35}$ \\
(c) Adaptive+$\Delta_t^2 x_t-\Delta_t^2\widetilde{x}_t$+Conv2D & \cellcolor{orange!24} $91.43{\scriptscriptstyle\pm0.67}$ & $82.16{\scriptscriptstyle\pm0.39}$ & $80.20{\scriptscriptstyle\pm0.22}$ & $84.61{\scriptscriptstyle\pm0.48}$ & $45.91{\scriptscriptstyle\pm0.66}$ & $76.86{\scriptscriptstyle\pm0.32}$ \\
(d) Adaptive+Errors\&$\Delta_t^2 x_t-\Delta_t^2\widetilde{x}_t$+Conv3D & $89.53{\scriptscriptstyle\pm0.33}$ & \cellcolor{cyan!24} $85.35{\scriptscriptstyle\pm0.31}$ & \cellcolor{cyan!24} $85.82{\scriptscriptstyle\pm0.18}$ & \cellcolor{cyan!24} $89.75{\scriptscriptstyle\pm0.06}$ & \cellcolor{orange!24} $60.62{\scriptscriptstyle\pm0.66}$ & \cellcolor{orange!24} $82.21{\scriptscriptstyle\pm0.26}$ \\
\midrule
(Ours) Adaptive+Errors+Conv3D & \cellcolor{cyan!24} $93.72{\scriptscriptstyle\pm2.27}$ & \cellcolor{orange!24} $82.89{\scriptscriptstyle\pm0.22}$ & \cellcolor{orange!24} $82.72{\scriptscriptstyle\pm1.99}$ & \cellcolor{orange!24} $87.95{\scriptscriptstyle\pm1.10}$ & \cellcolor{cyan!24} $65.03{\scriptscriptstyle\pm0.81}$ & \cellcolor{cyan!24} $82.46{\scriptscriptstyle\pm0.89}$ \\
\bottomrule
\end{tabular}
\end{adjustbox}
\end{table}

\begin{table}[H]
\centering
\caption{\revise{Robustness under video compression settings on HRFVideo. }}
\label{tab:postprocessing_hrfvideo}
\revise{\begin{minipage}{\textwidth}
\centering\footnotesize
\setlength{\tabcolsep}{2pt}
\renewcommand{\arraystretch}{1.12}
\begin{tabular*}{\textwidth}{@{\extracolsep{\fill}}llcccccc@{}}
\toprule
Level & Compression & Wan2.1 & Hailuo & Seaweed & Seedance & StepVideo & Avg. \\
\midrule
\multirow{3}{*}{Video} & H.264 CRF 18 & 98.02/92.73 & 93.70/82.43 & 94.03/82.76 & 95.39/87.25 & 89.11/62.75 & 94.05/81.58 \\
 & H.264 CRF 23 & 97.92/92.57 & 92.99/80.50 & 93.74/81.81 & 94.14/84.42 & 88.63/61.86 & 93.48/80.23 \\
 & H.264 CRF 28 & 97.65/92.12 & 91.55/78.53 & 92.98/80.41 & 93.91/81.93 & 87.73/60.56 & 92.76/78.71 \\
\midrule
\multirow{3}{*}{Frame} & JPEG & 98.46/94.19 & 95.22/85.70 & 95.27/85.60 & 96.39/89.08 & 91.25/65.81 & 95.32/84.08 \\
 & Downsample & 97.82/91.84 & 93.96/82.79 & 93.89/82.28 & 95.47/87.03 & 92.49/68.37 & 94.73/82.46 \\
 & JPEG + Downsample & 98.45/94.22 & 95.42/86.63 & 95.14/85.44 & 96.45/89.54 & 93.98/70.40 & 95.89/85.25 \\
 \midrule
\multicolumn{2}{@{}c}{No Compression} & 98.57/96.29 & 93.56/83.14 & 94.78/85.02 & 95.60/88.75 & 89.98/64.17 & 94.50/83.47 \\
\bottomrule
\end{tabular*}
\end{minipage}}
\end{table}

\begin{table}[H]
\centering
\caption{\revise{Detection performance under different temporal sampling rates on FakeParts. }}
\label{tab:temporal_sampling}
\revise{\begin{minipage}{\textwidth}
\centering\footnotesize
\setlength{\tabcolsep}{2pt}
\renewcommand{\arraystretch}{1.12}
\begin{tabular*}{\textwidth}{@{\extracolsep{\fill}}lccccc@{}}
\toprule
Sampling rate & CogVideoX & LTX-Video & AllegroAI & Veo2 & Avg. \\
\midrule
2 fps & 99.11/99.11 & 94.32/93.17 & 99.18/98.87 & 96.77/96.47 & 97.35/96.91 \\
4 fps & 99.36/99.38 & 94.64/93.67 & 98.57/98.40 & 94.59/93.88 & 96.79/96.33 \\
8 fps & 98.39/98.52 & 95.70/95.03 & 98.87/98.78 & 95.24/94.51 & 97.05/96.71 \\
\bottomrule
\end{tabular*}
\par\medskip
\begin{tabular*}{\textwidth}{@{\extracolsep{\fill}}lcccccc@{}}
\end{tabular*}
\end{minipage}}
\end{table}

\begin{table}[H]
\centering
\caption{\revise{Average inference cost per video on FakeParts with an A100-80GB GPU. TFLOPs denotes trillions of floating-point operations per video during inference.}}
\label{tab:inference_cost}
\begingroup\color{black}\small
\setlength{\tabcolsep}{6pt}
\renewcommand{\arraystretch}{0.95}
\begin{tabular}{@{}lrr@{}}
\toprule
Method & TFLOPs & Latency (s) \\
\midrule
D3 & 0.564 & 0.0167 \\
STALL & 2.011 & 0.0313 \\
Over-Coherence & 2.592 & 0.0295 \\
ReConFuse & 6.028 & 0.0748 \\
DIVID-CNN & 2,240.074 & 25.9296 \\
DIVID-LSTM & 2,241.113 & 25.9251 \\
G2VD-CLIP & 0.281 & 0.0083 \\
DeMamba & 0.295 & 0.0234 \\
Ivy-xDetector & 3.635 & 7.8562 \\
DeepTraceReward & 6.525 & 0.7523 \\
MM-Det & 12.257 & 1.4049 \\
\midrule
\textbf{Ours} & 6.309 & 0.1170 \\
\bottomrule
\end{tabular}
\endgroup
\end{table}

\end{document}